\documentclass{article} 
\usepackage{iclr2025_conference,times}

\usepackage[pagebackref,breaklinks,colorlinks,citecolor=cvprblue]{hyperref}
\usepackage{url}
\usepackage{booktabs}
\usepackage{multirow}
\usepackage{graphicx}
\usepackage{xspace}

\usepackage{mathtools}
\usepackage{adjustbox}
\usepackage{array}
\usepackage{colortbl}
\usepackage{float,wrapfig}
\usepackage{hhline}
\usepackage{subcaption} 
\usepackage{enumitem}

\usepackage{tikz}
\usetikzlibrary{tikzmark}

\usepackage{wrapfig}

\def\tablecite#1{[\citenum{#1}]}
\newcolumntype{g}{>{\columncolor{mitblue}}c}

\newcolumntype{i}{>{\columncolor{gray}}c}

\definecolor{cvprblue}{rgb}{0.21,0.49,0.74}
\definecolor{mitblue}{rgb}{0.88,0.95,0.96}
\definecolor{gold}{rgb}{0.75,0.6,0.12}
\colorlet{shadecolor}{gray!40}

\usepackage{amsmath,amsfonts,bm}

\def\eqref#1{equation~\ref{#1}}

\def\1{\bm{1}}

\DeclareMathAlphabet{\mathsfit}{\encodingdefault}{\sfdefault}{m}{sl}
\SetMathAlphabet{\mathsfit}{bold}{\encodingdefault}{\sfdefault}{bx}{n}

\def\modelfull{Deep Compression Autoencoder\xspace}
\def\modelshort{DC-AE\xspace}

\makeatletter
\def\blfootnote#1{\xdef\@thefnmark{}\@footnotetext{\scriptsize #1}}
\makeatother

\title{\modelfull for \\ Efficient High-Resolution Diffusion Models}

\author{
    Junyu Chen$^{1,2*}$, \space\space\space
    Han Cai$^{3*\dag}$, \space\space\space
    Junsong Chen$^3$, \space\space\space
    Enze Xie$^3$, \\
    \textbf{
    Shang Yang$^1$, \space\space\space
    Haotian Tang$^1$, \space\space\space
    Muyang Li$^1$, \space\space\space
    Yao Lu$^3$, \space\space\space
    Song Han$^{1,3}$
    } \\
    \:$^{1}$\normalfont{MIT} \quad
    $^{2}$\normalfont{Tsinghua University} \quad
    $^{3}$\normalfont{NVIDIA} \\
    \:\url{https://github.com/mit-han-lab/efficientvit}
}

\definecolor{mydarkred}{rgb}{0.8,0.02,0.02}

\iclrfinalcopy 
\begin{document}

\maketitle

\blfootnote{$^*$Equal contribution. Junyu Chen is an intern at MIT during this work.}

\blfootnote{$^\dag$Project lead. Correspondence to: \texttt{hcai@nvidia.com, songhan@mit.edu}.}

\begin{abstract}
We present \modelfull (\modelshort), a new family of autoencoders for accelerating high-resolution diffusion models. Existing autoencoders have demonstrated impressive results at a moderate spatial compression ratio (e.g., 8$\times$), but fail to maintain satisfactory reconstruction accuracy for high spatial compression ratios (e.g., 64$\times$). We address this challenge by introducing two key techniques: (1) \textbf{Residual Autoencoding}, where we design our models to learn residuals based on the space-to-channel transformed features to alleviate the optimization difficulty of high spatial-compression autoencoders; (2) \textbf{Decoupled High-Resolution Adaptation}, an efficient decoupled three-phase training strategy for mitigating the generalization penalty of high spatial-compression autoencoders. With these designs, we improve the autoencoder's spatial compression ratio up to 128 while maintaining the reconstruction quality. Applying our \modelshort to latent diffusion models, we achieve significant speedup without accuracy drop. For example, on ImageNet $512 \times 512$, our \modelshort provides \textbf{19.1$\times$} inference speedup and \textbf{17.9$\times$} training speedup on H100 GPU for UViT-H while achieving a better FID, compared with the widely used SD-VAE-f8 autoencoder.
\end{abstract}

\section{Introduction}
\label{sec:intro}

Latent diffusion models \citep{rombach2022high} have emerged as a leading framework and demonstrated great success in image synthesis \citep{flux2024,esser2024scaling}. They employ an autoencoder to project the images to the latent space to reduce the cost of diffusion models. For example, the predominantly adopted solution in current latent diffusion models \citep{rombach2022high,flux2024,esser2024scaling,chenpixart,chen2024pixart} is to use an autoencoder with a spatial compression ratio of 8 (denoted as f8), which converts 
images of spatial size $H\times W$ to latent features of spatial size $\frac{H}{8} \times \frac{W}{8}$. This spatial compression ratio is satisfactory for low-resolution image synthesis (e.g., $256 \times 256$). However, for high-resolution image synthesis (e.g., $1024 \times 1024$), further increasing the spatial compression ratio is critical, especially for diffusion transformer models \citep{peebles2023scalable, bao2023all} that have quadratic computational complexity to the number of tokens.

The current common practice for further reducing the spatial size is downsampling on the diffusion model side. For example, in diffusion transformer models \citep{peebles2023scalable, bao2023all}, this is achieved by using a patch embedding layer with patch size $p$ that compresses the latent features to $\frac{H}{8p} \times \frac{W}{8p}$ tokens. In contrast, little effort has been made on the autoencoder side. The main bottleneck hindering the employment of high spatial-compression autoencoders is the reconstruction accuracy drop. For example, Figure~\ref{fig:figure1_results} (a) shows the reconstruction results of SD-VAE \citep{rombach2022high} on ImageNet $256 \times 256$ with different spatial compression ratios. We can see that the rFID (reconstruction FID) degrades from 0.90 to 28.3 if switching from f8 to f64. 

This work presents \textbf{\modelfull (\modelshort)}, a new family of high spatial-compression autoencoders for efficient high-resolution image synthesis. By analyzing the underlying source of the accuracy degradation between high spatial-compression and low spatial-compression autoencoders, we find high spatial-compression autoencoders are more difficult to optimize (Section~\ref{sec:motivation}) and suffer from the generalization penalty across resolutions (Figure~\ref{fig:method_motivation} b). To this end, we introduce two key techniques to address these two challenges. First, we propose \textbf{Residual Autoencoding} (Figure~\ref{fig:method_arch}) to alleviate the optimization difficulty of high spatial-compression autoencoders. It introduces extra non-parametric shortcuts to the autoencoder to let the neural network modules learn residuals based on the space-to-channel operation. Second, we propose \textbf{Decoupled High-Resolution Adaptation} (Figure~\ref{fig:method_training_pipeline}) to tackle the other challenge. It introduces a high-resolution latent adaptation phase and a low-resolution local refinement phase to avoid the generalization penalty while maintaining a low training cost. 

With these techniques, we increase the spatial compression ratio of autoencoders to 32, 64, and 128 while maintaining good reconstruction accuracy (Table~\ref{tab:ae_main}). The diffusion models can fully focus on the denoising task with our \modelshort taking over the whole token compression task, which delivers better image generation results than prior approaches (Table~\ref{tab:diffusion_imagenet_main}). For example, replacing SD-VAE-f8 with our \modelshort-f64, we achieve \textbf{17.9$\times$} higher H100 training throughput and \textbf{19.1$\times$} higher H100 inference throughput on UViT-H \citep{bao2023all} while improving the ImageNet $512 \times 512$ FID from 3.55 to 3.01.  
We summarize our contributions as follows:

\vspace{-5pt}
\begin{itemize}[leftmargin=*]
\item We analyze the challenges of increasing the spatial compression ratio of autoencoders and provide insights into how to address these challenges.  
\item We propose Residual Autoencoding and Decoupled High-Resolution Adaptation that effectively improve the reconstruction accuracy of high spatial-compression autoencoders, making their reconstruction accuracy feasible for use in latent diffusion models. 
\item We build \modelshort, a new family of autoencoders based on our techniques. It delivers significant training and inference speedup for diffusion models compared with prior autoencoders. 
\end{itemize}

\begin{figure}[t]
    \centering
    \includegraphics[width=1\linewidth]{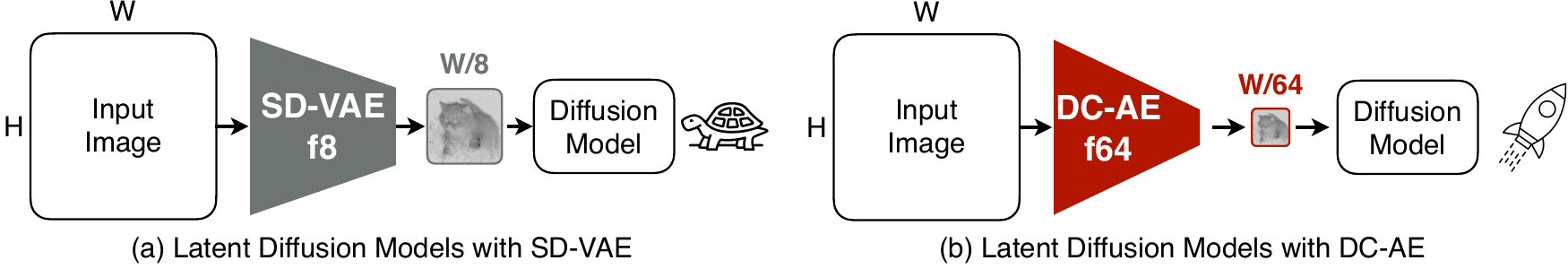}
    \vspace{-15pt}
    \caption{\modelshort accelerates diffusion models by increasing autoencoder's spatial compression ratio.}
    \vspace{-10pt}
    \label{fig:figure1}
\end{figure}

\begin{figure}[t]
    \centering
    \includegraphics[width=1\linewidth]{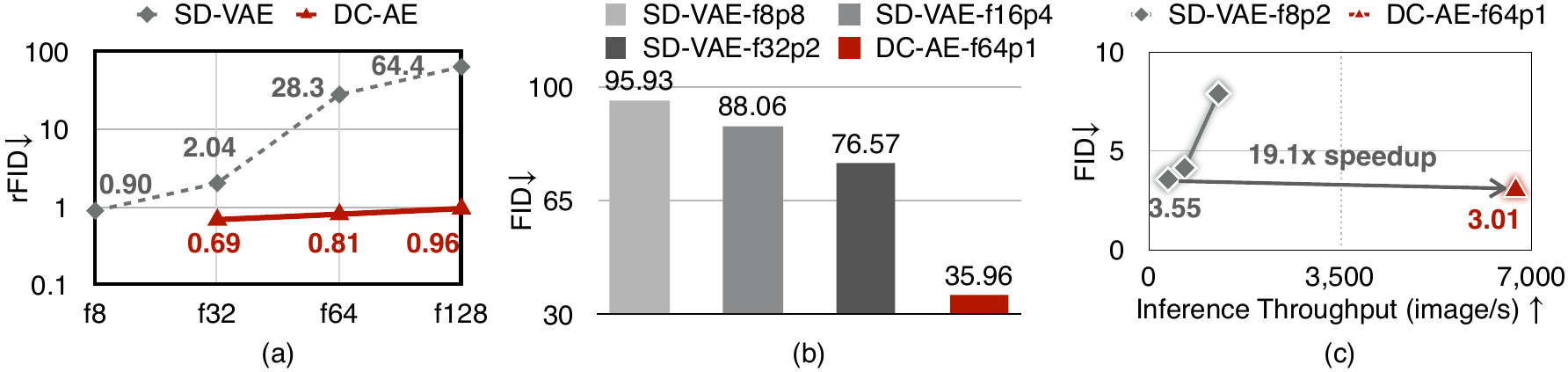}
    \vspace{-15pt}
    \caption{\textbf{(a) Image Reconstruction Results on ImageNet 256$\times$256.} f denotes the spatial compression ratio. When the spatial compression ratio increases, SD-VAE has a significant reconstruction accuracy drop (higher rFID) while \modelshort does not have this issue. \textbf{(b) ImageNet 512$\times$512 Image Generation Results on UViT-S with Various Autoencoders.} p denotes the patch size. Shifting the token compression task to the autoencoder enables the diffusion model to focus more on the denoising task, leading to better FID. \textbf{(c) Comparison to SD-VAE-f8 on ImageNet 512$\times$512 with UViT Variants.} \modelshort-f64p1 provides 19.1$\times$ higher inference throughput and 0.54 better ImageNet FID than SD-VAE-f8p2 on UViT-H.}
    \vspace{-10pt}
    \label{fig:figure1_results}
\end{figure}

\section{Related Work}
\label{sec:related}

\paragraph{Autoencoder for Diffusion Models.}
Training and evaluating diffusion models directly in high-resolution pixel space results in prohibitive computational costs. To address this issue, \cite{rombach2022high} proposes latent diffusion models that operate in a compressed latent space produced by pretrained autoencoders. The proposed autoencoder with $8\times$ spatial compression ratio and $4$ latent channels has been widely adopted in subsequent works \citep{peebles2023scalable,bao2023all}. Since then, follow-up works mainly focus on enhancing the reconstruction accuracy of the f8 autoencoder by increasing the number of latent channels \citep{esser2024scaling,dai2023emu,flux2024}. Additionally, to improve the reconstruction quality, \cite{zhu2023designing} leverages a heavier decoder and incorporates task-specific priors. In contrast to prior works, our work focuses on an orthogonal direction, increasing the spatial compression ratio of the autoencoders (e.g., f64). To the best of our knowledge, our work is the first study in this critical but underexplored direction. 

\vspace{-5pt}
\paragraph{Diffusion Model Acceleration.}
Diffusion models have been widely used for image generation and showed impressive results \citep{flux2024,esser2024scaling}. However, diffusion models are computationally intensive, motivating many works to accelerate diffusion models. One representative strategy is reducing the number of inference sampling steps by training-free few-step samplers \citep{songdenoising, lu2022dpm, lu2022dpm2, zheng2023dpm, zhangfast, zhanggddim, zhao2024unipc, shih2024parallel, tang2024accelerating} or distilling-based methods \citep{meng2023distillation, salimans2022progressive, yin2024one, yin2024improved, song2023consistency, luo2023latent, liu2023instaflow}. Another representative strategy is model compression by leveraging sparsity \citep{li2022efficient, ma2024deepcache} or quantization \citep{he2024ptqd, fang2024structural, li2023q, zhao2024vidit}. Designing efficient diffusion model architectures \citep{li2024snapfusion,liu2024linfusion,cai2024condition} or inference systems \citep{li2024distrifusion, wang2024pipefusion} is also an effective approach for boosting efficiency. In addition, improving the data quality \citep{chenpixart,chen2024pixart} can boost the training efficiency of diffusion models. 

All these works focus on diffusion models while the autoencoder remains the same. Our work opens up a new direction for accelerating diffusion models, which can benefit both training and inference.
\begin{figure}[t]
    \centering
    \includegraphics[width=1\linewidth]{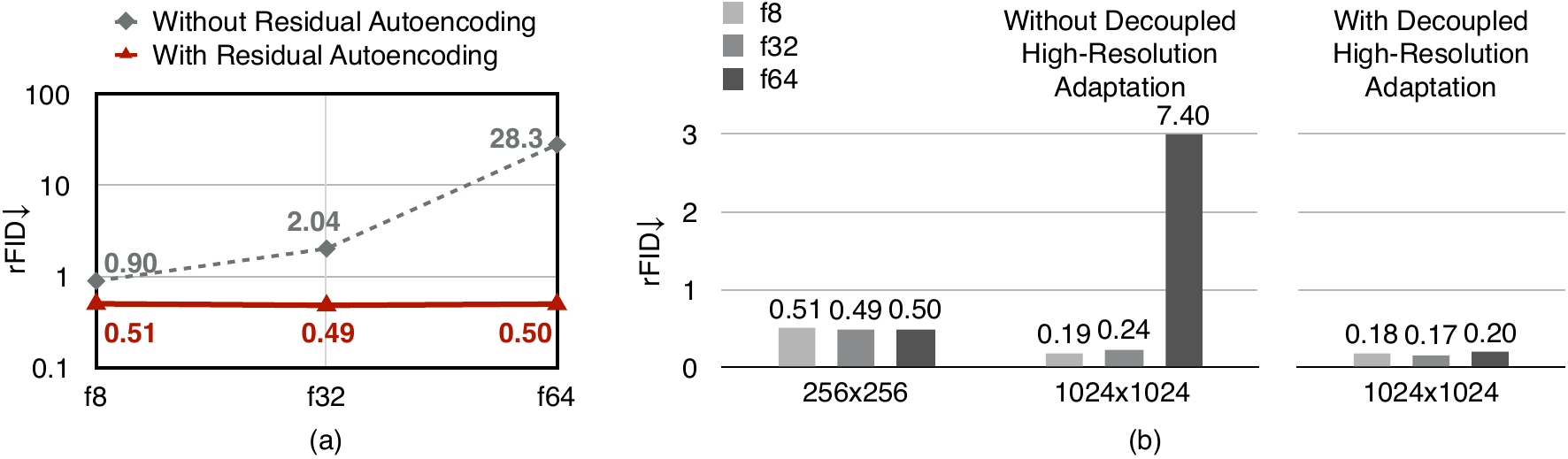}
    \vspace{-15pt}
    \caption{(a) High spatial-compression autoencoders are more difficult to optimize. Even with the same latent shape and stronger learning capacity, it still cannot match the f8 autoencoder's rFID. (b) High spatial-compression autoencoders suffer from significant reconstruction accuracy drops when generalizing from low-resolution to high-resolution. }
    \vspace{-10pt}
    \label{fig:method_motivation}
\end{figure}

\section{Method}
\label{sec:method}
\vspace{-5pt}
In this section, we first analyze why existing high spatial-compression autoencoders (e.g., SD-VAE-f64) fail to match the accuracy of low spatial-compression autoencoders (e.g., SD-VAE-f8). Then we introduce our \modelfull (\modelshort) with \emph{Residual Autoencoding} and \emph{Decoupled High-Resolution Adaptation} to close the accuracy gap. Finally, we discuss the applications of our \modelshort to latent diffusion models. 

\subsection{Motivation}\label{sec:motivation}
We conduct ablation study experiments to get insights into the underlying source of the accuracy gap between high spatial-compression and low spatial-compression autoencoders. Specifically, we consider three settings with gradually increased spatial compression ratio, from f8 to f64. 

Each time the spatial compression ratio increases, we stack additional encoder and decoder stages upon the current autoencoder. In this way, high spatial-compression autoencoders contain low spatial-compression autoencoders as sub-networks and thus have higher learning capacity.

Additionally, we increase the latent channel number to maintain the same total latent size across different settings. We can then convert the latent to a higher spatial compression ratio one by applying a space-to-channel operation \citep{shi2016real}: $H \times W \times C \rightarrow \frac{H}{p} \times \frac{W}{p} \times p^2C$. 

We summarize the results in Figure~\ref{fig:method_motivation} (a, gray dash line). Even with the same total latent size and stronger learning capacity, we still observe degraded reconstruction accuracy when the spatial compression ratio increases. It demonstrates that \emph{the added encoder and decoder stages (consisting of multiple SD-VAE building blocks) work worse than a simple space-to-channel operation}. 

Based on this finding, we conjecture \emph{the accuracy gap comes from the model learning process: while we have good local optimums in the parameter space, the optimization difficulty hinders high spatial-compression autoencoders from reaching such local optimums.}

\begin{figure}[t]
    \centering
    \includegraphics[width=0.9\linewidth]{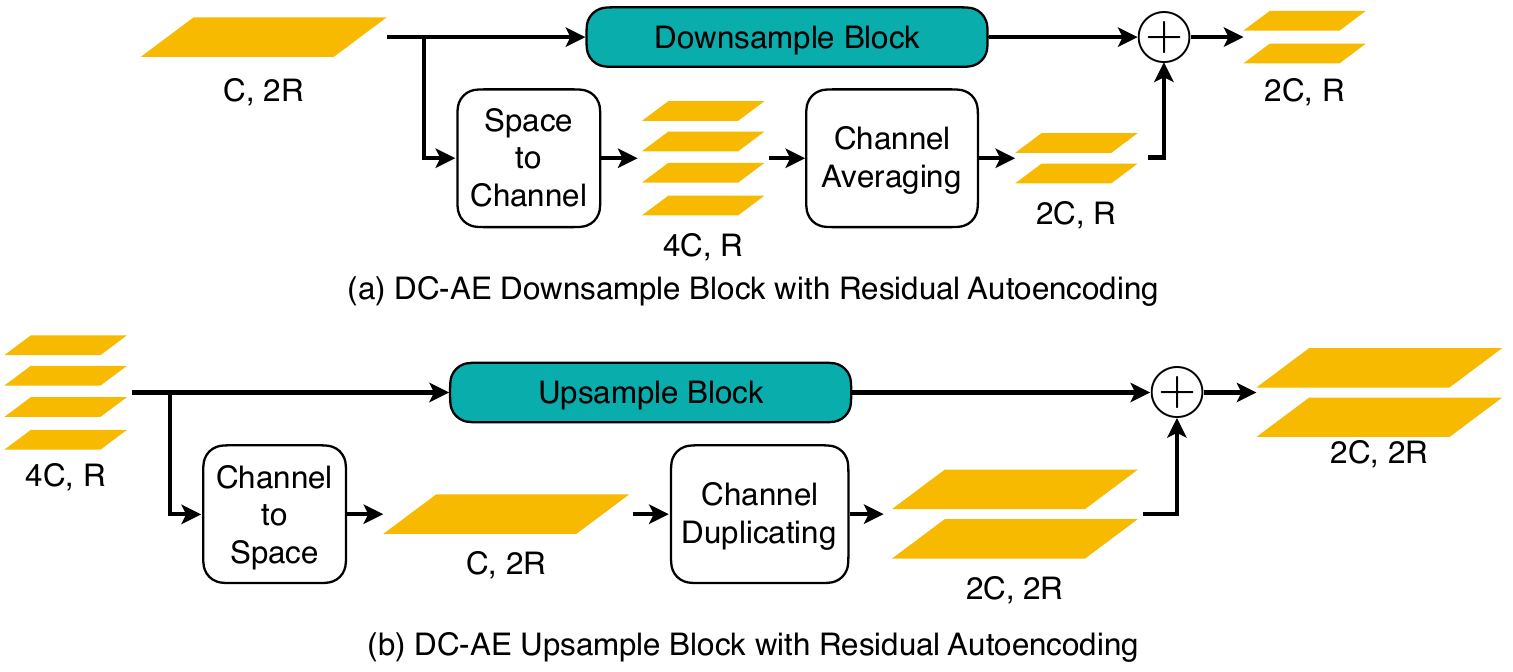}
    \vspace{-5pt}
    \caption{\textbf{Illustration of Residual Autoencoding.} It adds non-parametric shortcuts to let the neural network modules learn residuals based on the space-to-channel operation. `C' denotes the number of channels. `R' denotes the image size. }
    \vspace{-5pt}
    \label{fig:method_arch}
\end{figure}

\subsection{\modelfull}\label{sec:main_model}
\paragraph{Residual Autoencoding.} Motivated by the analysis, we introduce Residual Autoencoding to address the accuracy gap. The general idea is depicted in Figure~\ref{fig:method_arch}. The core difference from the conventional design is that we explicitly let neural network modules learn the downsample residuals based on the space-to-channel operation to alleviate the optimization difficulty. Different from ResNet \citep{he2016deep}, the residual here is not identity mapping, but space-to-channel mapping.

In practice, this is implemented by adding extra non-parametric shortcuts on the encoder's downsample blocks and decoder's upsample blocks. Specifically, for the downsample block, the non-parametric shortcut is a space-to-channel operation followed by a non-parametric channel averaging operation to match the channel number. For example, assuming the downsample block's input feature map shape is $H \times W \times C$ and its output feature map shape is $\frac{H}{2} \times \frac{W}{2} \times 2C$, then the added shortcut is:

\vspace{-10pt}
{\small
\begin{align}
    H \times W \times C & \xrightarrow{\text{space-to-channel}} \frac{H}{2} \times \frac{W}{2} \times 4C \nonumber \\
    & \underbrace{\xrightarrow{\text{split into two groups}} [\frac{H}{2} \times \frac{W}{2} \times 2C, \frac{H}{2} \times \frac{W}{2} \times 2C] \xrightarrow{\text{average}} \frac{H}{2} \times \frac{W}{2} \times 2C.}_{\text{channel averaging}} \nonumber
\end{align}
}
\vspace{-10pt}

Accordingly, for the upsample block, the non-parametric shortcut is a channel-to-space operation followed by a non-parametric channel duplicating operation:

\vspace{-10pt}
{\small
\begin{align}
    \frac{H}{2} \times \frac{W}{2} \times 2C & \xrightarrow{\text{channel-to-space}}  \nonumber H \times W \times \frac{C}{2} \\
    & \underbrace{\xrightarrow{\text{duplicate}} [H \times W \times \frac{C}{2}, H \times W \times \frac{C}{2}] \xrightarrow{\text{concat}} H \times W \times C.}_{\text{channel duplicating}} \nonumber
\end{align}
}
\vspace{-10pt}

In addition to the downsample and upsample blocks, we also change the middle stage design following the same principle (Figure~\ref{fig:dc_ae_detailed_arch} b, right). 

Figure~\ref{fig:method_motivation} (a) shows the comparison with and without our Residual Autoencoding on ImageNet $256 \times 256$. We can see that Residual Autoencoding effectively improves the reconstruction accuracy of high spatial-compression autoencoders. 

\begin{figure}[t]
    \centering
    \includegraphics[width=0.95\linewidth]{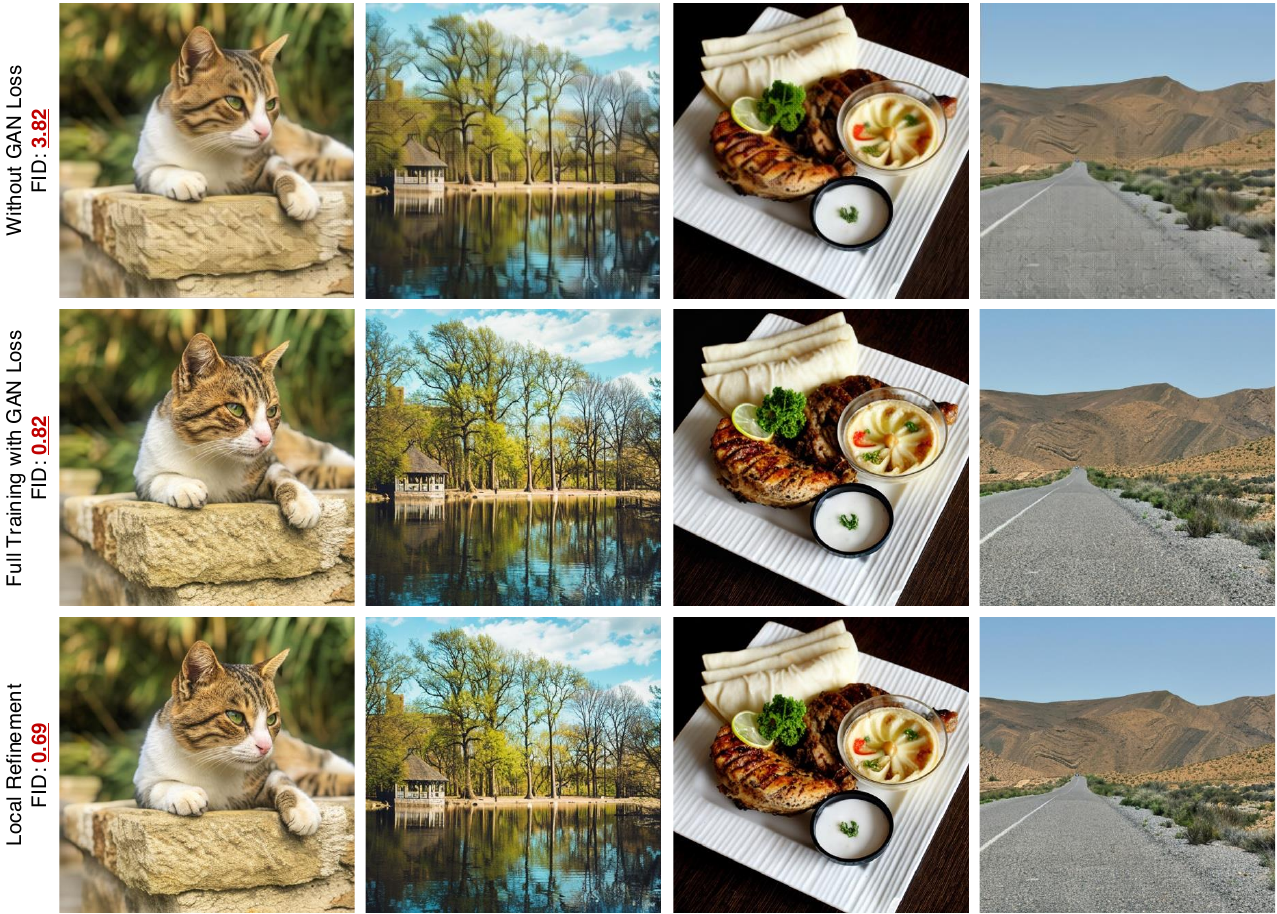}
    \caption{Autoencoder already learns to reconstruct content and semantics without GAN loss, while GAN loss improves local details and removes local artifacts. We replace the GAN loss full training with lightweight local refinement training which achieves the same goal and has lower training cost.}
    \vspace{-5pt}
    \label{fig:method_gan}
\end{figure}
\begin{figure}[t]
    \centering
    \includegraphics[width=0.95\linewidth]{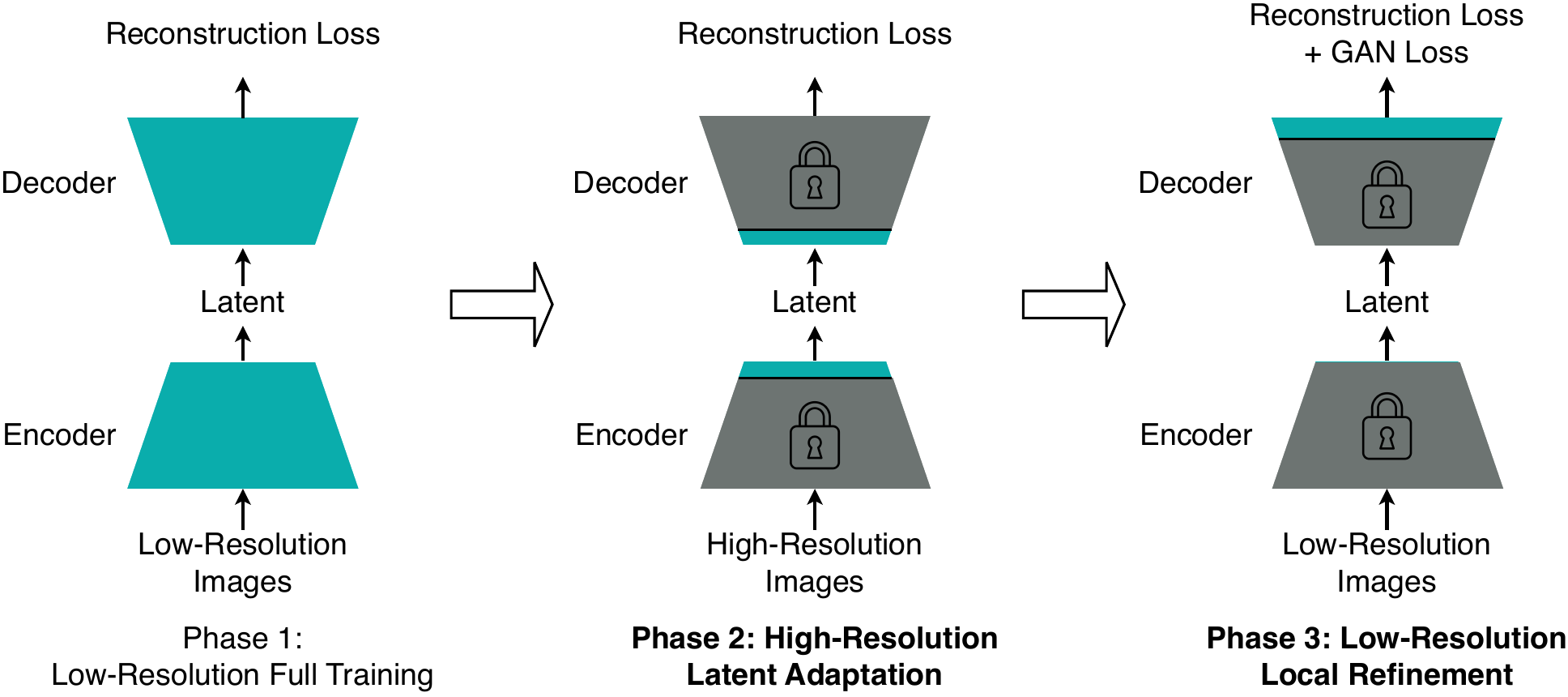}
    \caption{\textbf{Illustration of Decoupled High-Resolution Adaptation.}}
    \vspace{-15pt}
    \label{fig:method_training_pipeline}
\end{figure}

\paragraph{Decoupled High-Resolution Adaptation.} Residual Autoencoding alone can address the accuracy gap when handling low-resolution images. However, when extending it to high-resolution images, we find it not sufficient. Due to the large cost of high-resolution training, the common practice for high-resolution diffusion models is directly using autoencoders trained on low-resolution images (e.g., $256\times256$) \citep{chenpixart,chen2024pixart}. This strategy works well for low spatial-compression autoencoders. However, high spatial-compression autoencoders suffer from a significant accuracy drop. For example, in Figure~\ref{fig:method_motivation} (b), we can see that f64 autoencoder's rFID degrades from 0.50 to 7.40 when generalizing from $256\times256$ to $1024\times1024$. In contrast, the f8 autoencoder's rFID improves from 0.51 to 0.19 under the same setting. Additionally, we also find this issue more severe when using a higher spatial compression ratio. In this work, we refer to this phenomenon as the \emph{generalization penalty of high spatial-compression autoencoders}. A straightforward solution to address this issue is conducting training on high-resolution images. However, it suffers from a large training cost and unstable high-resolution GAN loss training. 

We introduce Decoupled High-Resolution Adaptation to tackle this challenge. Figure~\ref{fig:method_training_pipeline} demonstrates the detailed training pipeline. Compared with the conventional single-phase training strategy \citep{rombach2022high}, our Decoupled High-Resolution Adaptation has two key differences. 

First, we decouple the GAN loss training from the full model training and introduce a dedicated local refinement phase for the GAN loss training. In the local refinement phase (Figure~\ref{fig:method_training_pipeline}, phase 3), we only tune the head layers of the decoder while freezing all the other layers. The intuition of this design is based on the finding that the reconstruction loss alone is sufficient for learning to reconstruct the content and semantics. Meanwhile, the GAN loss mainly improves local details and removes local artifacts (Figure~\ref{fig:method_gan}). Achieving the same goal of local refinement, only tuning the decoder's head layers has a lower training cost and delivers better accuracy than the full training. 

Moreover, the decoupling prevents the GAN loss training from changing the latent space. This approach enables us to conduct the local refinement phase on low-resolution images without worrying about the generalization penalty. This further reduces the training cost of phase 3 and avoids the highly unstable high-resolution GAN loss training. 

Second, we introduce an additional high-resolution latent adaptation phase (Figure~\ref{fig:method_training_pipeline}, phase 2) that tunes the middle layers (i.e., encoder's head layers and decoder's input layers) to adapt the latent space for alleviating the generalization penalty. In our experiments, we find only tuning middle layers is sufficient for addressing this issue (Figure~\ref{fig:method_motivation} b) while having a lower training cost than high-resolution full training (memory cost: 153.98 GB $\rightarrow$ 67.81 GB)\footnote{Assuming the input resolution is $1024 \times 1024$ and the batch size is 12.} \citep{cai2020tinytl}. 

\begin{table}[t]
\small\centering\setlength{\tabcolsep}{4pt}
\begin{tabular}{l | c | c | c | c c }
\toprule
\multicolumn{5}{l}{\textbf{ImageNet 512$\times$512 (Class-Conditional)}} \\
\midrule
Diffusion Model & Autoencoder & Patch Size & \#Tokens & FID (w/o CFG) $\downarrow$ &  FID (w/ CFG) $\downarrow$ \\
\midrule
& SD-VAE-f8       & 8 & 64 & 125.08 & 95.93 \\
& SD-VAE-f16      & 4 & 64 & 115.32 & 88.06 \\
& SD-VAE-f32      & 2 & 64 & 107.33 & 76.57 \\
\cmidrule{2-6}
\multirow{-4}{*}{UViT-S \tablecite{bao2023all}} 
& \modelshort-f64 & 1 & 64 & \textbf{67.30} & \textbf{35.96} \\
\bottomrule
\end{tabular}
\caption{\textbf{Ablation Study on Patch Size and Autoencoder's Spatial Compression Ratio.}}
\label{tab:diffusion_ablation_vae_patchsize}
\end{table}

\subsection{Application to Latent Diffusion Models}
Applying our \modelshort to latent diffusion models is straightforward. The only hyperparameter to change is the patch size \citep{peebles2023scalable}. For diffusion transformer models \citep{peebles2023scalable,bao2023all}, increasing the patch size $p$ is the common approach for reducing the number of tokens. It is equivalent to first applying the space-to-channel operation to reduce the spatial size of the given latent by $p \times$ and then using the transformer model with a patch size of 1. 

Since combining a low spatial-compression autoencoder (e.g., f8) with the space-to-channel operation can also achieve a high spatial compression ratio, a natural question is how it compares with directly reaching the target spatial compression ratio with \modelshort. 

We conduct ablation study experiments and summarize the results in Table~\ref{tab:diffusion_ablation_vae_patchsize}. We can see that directly reaching the target spatial compression ratio with the autoencoder gives the best results among all settings. In addition, we also find that shifting the spatial compression ratio from the diffusion model to the autoencoder consistently leads to better FID. 

\section{Experiments}
\label{sec:exp}

\begin{table}[t]
\small\centering\setlength{\tabcolsep}{3pt}
\begin{tabular}{l | c | g | g g g g }
\toprule
\rowcolor{white} \textbf{ImageNet 256$\times$256} & Latent Shape & Autoencoder & rFID $\downarrow$ & PSNR $\uparrow$ & SSIM $\uparrow$ & LPIPS $\downarrow$ \\
\midrule
\rowcolor{white} \multirow{2}{*}{f32c32} & \multirow{2}{*}{8$\times$8$\times$32} 
 & SD-VAE \tablecite{rombach2022high} & 2.64 & 22.13 & 0.59 & 0.117 \\
 & & \modelshort                      & \textbf{0.69} & \textbf{23.85} & \textbf{0.66} & \textbf{0.082} \\
\midrule
\rowcolor{white} \multirow{2}{*}{f64c128} & \multirow{2}{*}{4$\times$4$\times$128}
 & SD-VAE \tablecite{rombach2022high} & 26.65 & 18.07 & 0.41 & 0.283 \\
 & & \modelshort                      & \textbf{0.81} & \textbf{23.60} & \textbf{0.65} & \textbf{0.087} \\
\bottomrule
\toprule
\rowcolor{white} \textbf{ImageNet 512$\times$512} & Latent Shape & Autoencoder & rFID $\downarrow$ & PSNR $\uparrow$ & SSIM $\uparrow$ & LPIPS $\downarrow$ \\
 \midrule
\rowcolor{white} \multirow{2}{*}{f64c128} & \multirow{2}{*}{8$\times$8$\times$128}
 & SD-VAE \tablecite{rombach2022high} & 16.84 & 19.49 & 0.48 & 0.282 \\
 & & \modelshort                      & \textbf{0.22} & \textbf{26.15} & \textbf{0.71} & \textbf{0.080} \\
\midrule
\rowcolor{white} \multirow{2}{*}{f128c512} & \multirow{2}{*}{4$\times$4$\times$512}
 & SD-VAE \tablecite{rombach2022high} & 100.74 & 15.90 & 0.40 & 0.531 \\
 & & \modelshort                      & \textbf{0.23} & \textbf{25.73} & \textbf{0.70} & \textbf{0.084} \\
\bottomrule
\toprule
\rowcolor{white} \textbf{FFHQ 1024$\times$1024} & Latent Shape & Autoencoder & rFID $\downarrow$ & PSNR $\uparrow$ & SSIM $\uparrow$ & LPIPS $\downarrow$ \\
\midrule
\rowcolor{white} \multirow{2}{*}{f64c128} & \multirow{2}{*}{16$\times$16$\times$128}
 & SD-VAE \tablecite{rombach2022high} & 6.62 & 24.55 & 0.68 & 0.237 \\
 & & \modelshort                      & \textbf{0.23} & \textbf{31.04} & \textbf{0.83} & \textbf{0.061} \\
 \midrule
\rowcolor{white} \multirow{2}{*}{f128c512} & \multirow{2}{*}{8$\times$8$\times$512}
 & SD-VAE \tablecite{rombach2022high} & 179.71 & 18.11 & 0.63 & 0.585 \\
 & & \modelshort                      &  \textbf{0.41} & \textbf{31.18} & \textbf{0.83} & \textbf{0.062} \\
\bottomrule
\toprule
\rowcolor{white} \textbf{MapillaryVistas 2048$\times$2048} & Latent Shape & Autoencoder & rFID $\downarrow$ & PSNR $\uparrow$ & SSIM $\uparrow$ & LPIPS $\downarrow$ \\ 
 \midrule
\rowcolor{white} \multirow{2}{*}{f64c128} & \multirow{2}{*}{32$\times$32$\times$128}
 & SD-VAE \tablecite{rombach2022high} & 7.55 & 22.37 & 0.68 & 0.262 \\
 & & \modelshort                      & \textbf{0.36} & \textbf{29.57} & \textbf{0.84} & \textbf{0.075} \\
 \midrule
\rowcolor{white} \multirow{2}{*}{f128c512} & \multirow{2}{*}{16$\times$16$\times$512} 
 & SD-VAE \tablecite{rombach2022high} & 152.09 & 17.82 & 0.67 & 0.594 \\
 & & \modelshort                      &  \textbf{0.38} & \textbf{29.70} & \textbf{0.84} & \textbf{0.074} \\
\bottomrule
\end{tabular}
\vspace{-5pt}
\caption{\textbf{Image Reconstruction Results.}}
\vspace{-10pt}
\label{tab:ae_main}
\end{table}

\subsection{Setups}

\paragraph{Implementation Details.} We use a mixture of datasets to train autoencoders (baselines and \modelshort), containing ImageNet \citep{deng2009imagenet}, SAM \citep{kirillov2023segment}, MapillaryVistas \citep{neuhold2017mapillary}, and FFHQ \citep{karras2019style}. For ImageNet experiments, we exclusively use the ImageNet training split to train autoencoders and diffusion models. The model architecture is similar to SD-VAE \citep{rombach2022high} except for our new designs discussed in Section~\ref{sec:main_model}. In addition, we use the original autoencoders instead of the variational autoencoders for our models, as they perform the same in our experiments and the original autoencoders are simpler. We also replace transformer blocks with EfficientViT blocks \citep{cai2023efficientvit} to make autoencoders more friendly for handling high-resolution images while maintaining similar accuracy. 

For image generation experiments, we apply autoencoders to diffusion transformer models including DiT \citep{peebles2023scalable} and UViT \citep{bao2023all}. We follow the same training settings as the original papers. Additionally, we build USiT by combining UViT \citep{bao2023all} with the SiT sampler \citep{ma2024sit}. The SiT and USiT models are trained for 500k iterations with batch size 1024.
We consider three settings with different resolutions, including ImageNet \citep{deng2009imagenet} for $512 \times 512$ generation, FFHQ \citep{karras2019style} and MJHQ \citep{li2024playground} for $1024 \times 1024$ generation, and MapillaryVistas \citep{neuhold2017mapillary} for $2048 \times 2048$ generation. 

\begin{table}[t]
\small\centering\setlength{\tabcolsep}{0.75pt}
\begin{tabular}{l | g | g | g | g g | g | g | g g }
\toprule
\rowcolor{white} Diffusion & & Patch & & \multicolumn{2}{c|}{Throughput (image/s) $\uparrow$} & Latency & Memory & \multicolumn{2}{c}{FID $\downarrow$} \\
\rowcolor{white} Model & \multirow{-2}{*}{Autoencoder} & Size & \multirow{-2}{*}{NFE} & Training & Inference & (ms) $\downarrow$ & (GB) $\downarrow$ & w/o CFG & w/ CFG \\
\midrule
\midrule
\rowcolor{white} & Flux-VAE-f8 \tablecite{flux2024}                          & 2 & 250 &   54 &   0.83 & 7915 & 56.3 & 27.35 & 8.72 \\
\cmidrule{2-10}
\rowcolor{white} & Asym-VAE-f8 \tablecite{zhu2023designing}                  & 2 & 250 &   54 &   0.85 & 7686 & 56.2 & 11.39 & 2.97 \\
\rowcolor{white} &  SD-VAE-f8 \tablecite{rombach2022high}                    & 2 & 250 &   54 &   0.85 & 7686 & 56.2 & 12.03 & 3.04 \\
\cmidrule{2-10}
& \modelshort-f32                                                            & 1 & 250 &  \textbf{241} &  \textbf{4.03} & \textbf{1958} & \textbf{20.9} &  9.56 & 2.84 \\
\multirow{-5}{*}{DiT-XL \tablecite{peebles2023scalable}} 
& \:\:\modelshort-f32$^\ddag$                                                & 1 & 250 &  \textbf{241} &  \textbf{4.03} & \textbf{1958} & \textbf{20.9} &  \textbf{6.88} & \textbf{2.41} \\

\midrule
\midrule
\rowcolor{white} & Flux-VAE-f8 \tablecite{flux2024}                          & 2 & 30 &  55 &   5.82 & 913 & 54.2 & 30.91 & 12.63 \\
\cmidrule{2-10}
\rowcolor{white} & Asym-VAE-f8 \tablecite{zhu2023designing}                  & 2 & 30 &  55 &   5.85 & 914 & 54.1 & 11.36 &  3.51 \\
\rowcolor{white}  & SD-VAE-f8 \tablecite{rombach2022high}                    & 2 & 30 &   \,\,55\,\,\tikzmark{uvit_h_f8p2:a1} &   \,\,5.85\,\,\tikzmark{uvit_h_f8p2:a2} & 914 & 54.1 & 11.04 & \,\,3.55\,\,\tikzmark{uvit_h_f8p2:a3} \\
\cmidrule{2-10}
& \modelshort-f32                                                            & 1 & 30 &  247 &  27.03 & 246 & 18.6 &  \textbf{9.83} & \textbf{2.53} \\
& \modelshort-f64                                                            & 1 & 30 & \,\,\textbf{984}\,\,\tikzmark{uvit_h_f64p1:a1} &  \,\,\textbf{111.77}\,\,\tikzmark{uvit_h_f64p1:a2} & \textbf{104} & \textbf{10.6} & 13.96 & \,\,3.01\,\,\tikzmark{uvit_h_f64p1:a3} \\
\multirow{-6}{*}{UViT-H \tablecite{bao2023all}} & \:\:\modelshort-f64$^\dag$ & 1 & 30 & \textbf{984} & \textbf{111.77} & \textbf{105} & \textbf{10.6} & 12.26 & 2.66 \\

\midrule
\midrule
\rowcolor{white} & Asym-VAE-f8 \tablecite{zhu2023designing}                   & 2 & 30 &  27 & 2.62 & 2243 & OOM & 9.87 & 3.62 \\
\rowcolor{white}  & SD-VAE-f8 \tablecite{rombach2022high}                     & 2 & 30 &  27 & 2.62 & 2243 & OOM & 9.73 & 3.57 \\
\cmidrule{2-10}
& \modelshort-f32                                                             & 1 & 30 & 112 & 11.08 & 590 & 42.0 & 8.13 & 2.30 \\
& \modelshort-f64                                                             & 1 & 30 & \textbf{450} & \textbf{45.55} & \textbf{258} & \textbf{30.2} & 7.78 & 2.47 \\
\multirow{-5}{*}{UViT-2B \tablecite{bao2023all}} & \:\:\modelshort-f64$^\dag$ & 1 & 30 & \textbf{450} & \textbf{45.55} & \textbf{258} & \textbf{30.2} & \textbf{6.50} & \textbf{2.25} \\

\midrule
\midrule
\rowcolor{white} MAGVIT-v2 \tablecite{yu2023language} & - & - & - & - & - & - & - & 3.07 & 1.91 \\
\rowcolor{white} EDM2-XXL \tablecite{karras2024analyzing} & - & - & - & - & - & - & - & \textbf{1.91} & 1.81 \\
\rowcolor{white} MAR-L \tablecite{li2024autoregressive} & - & - & - & - & - & - & - & 2.74 & 1.73 \\
\midrule
SiT-XL \tablecite{ma2024sit}        & \modelshort-f32 & 1 & - & 241 & - & - & 20.9 & 7.47 & 2.41 \\
USiT-H                              & \modelshort-f32 & 1 & - & 247 & - & - & 18.6 & 3.80 & 1.89 \\
USiT-2B                             & \modelshort-f32 & 1 & - & 112 & - & - & 42.0 & 2.90 & \textbf{1.72} \\

\bottomrule
\end{tabular}
\begin{tikzpicture}[overlay, remember picture, shorten >=.5pt, shorten <=.5pt, transform canvas={yshift=.25\baselineskip}]

\draw [->, red] ({pic cs:uvit_h_f8p2:a1}) [bend left] to node [below right] (uvit_h_f8p2:t1) {\hspace{-2pt}\scriptsize \textbf{17.9$\times$}} ({pic cs:uvit_h_f64p1:a1});

\draw [->, red] ({pic cs:uvit_h_f8p2:a2}) [bend left] to node [below right] (uvit_h_f8p2:t2) {\hspace{-2pt}\scriptsize \textbf{19.1$\times$}} ({pic cs:uvit_h_f64p1:a2});

\draw [->, red] ({pic cs:uvit_h_f8p2:a3}) [bend left] to node [below right] (uvit_h_f8p2:t3) {\hspace{-2pt}\scriptsize \textbf{-0.54}} ({pic cs:uvit_h_f64p1:a3});

\end{tikzpicture}
\vspace{-5pt}
\caption{\textbf{Class-Conditional Image Generation Results on ImageNet 512$\times$512.} $^\dag$ represents the model is trained for 4$\times$ training iterations (i.e., 500K $\rightarrow$ 2,000K iterations). $^\ddag$ represents the model is trained with 4$\times$ batch size (i.e., 256 $\rightarrow$ 1024). `NFE' denotes the number of functional evaluations. The NFEs for SiT \citep{ma2024sit} and USiT models are left blank as they use an adaptive-step evaluation scheduler.}
\label{tab:diffusion_imagenet_main}
\end{table}

\begin{table}[t]
\small\centering\setlength{\tabcolsep}{1pt}
\begin{tabular}{l | g | g | g | g g | g | g | g g }
\toprule
\rowcolor{white} Diffusion & & Patch & & \multicolumn{2}{c|}{Throughput (image/s) $\uparrow$} & Latency & Memory & \multicolumn{2}{c}{MJHQ 512$\times$512} \\
\rowcolor{white} Model & \multirow{-2}{*}{Autoencoder} & Size & \multirow{-2}{*}{NFE} & Training & Inference & (ms) $\downarrow$ & (GB) $\downarrow$ & FID $\downarrow$ & CLIP Score $\uparrow$ \\
\midrule
\rowcolor{white} & SD-VAE-f8 \tablecite{rombach2022high} & 2 & 20 & 43 & 7.81 & 742 & 60.45 & 6.3 & 26.36 \\
\multirow{-2}{*}{PIXART-$\alpha$ \tablecite{chenpixart}} 
& \modelshort-f32                                        & 1 & 20 & \textbf{173} & \textbf{31.27} & \textbf{209} & \textbf{23.77} & \textbf{6.1} & \textbf{26.41} \\
\bottomrule
\end{tabular}
\vspace{-5pt}
\caption{\textbf{Text-to-Image Generation Results.}}
\vspace{-15pt}
\label{tab:diffusion_t2i_main}
\end{table}

\begin{figure}[t]
    \centering
    \includegraphics[width=0.95\linewidth]{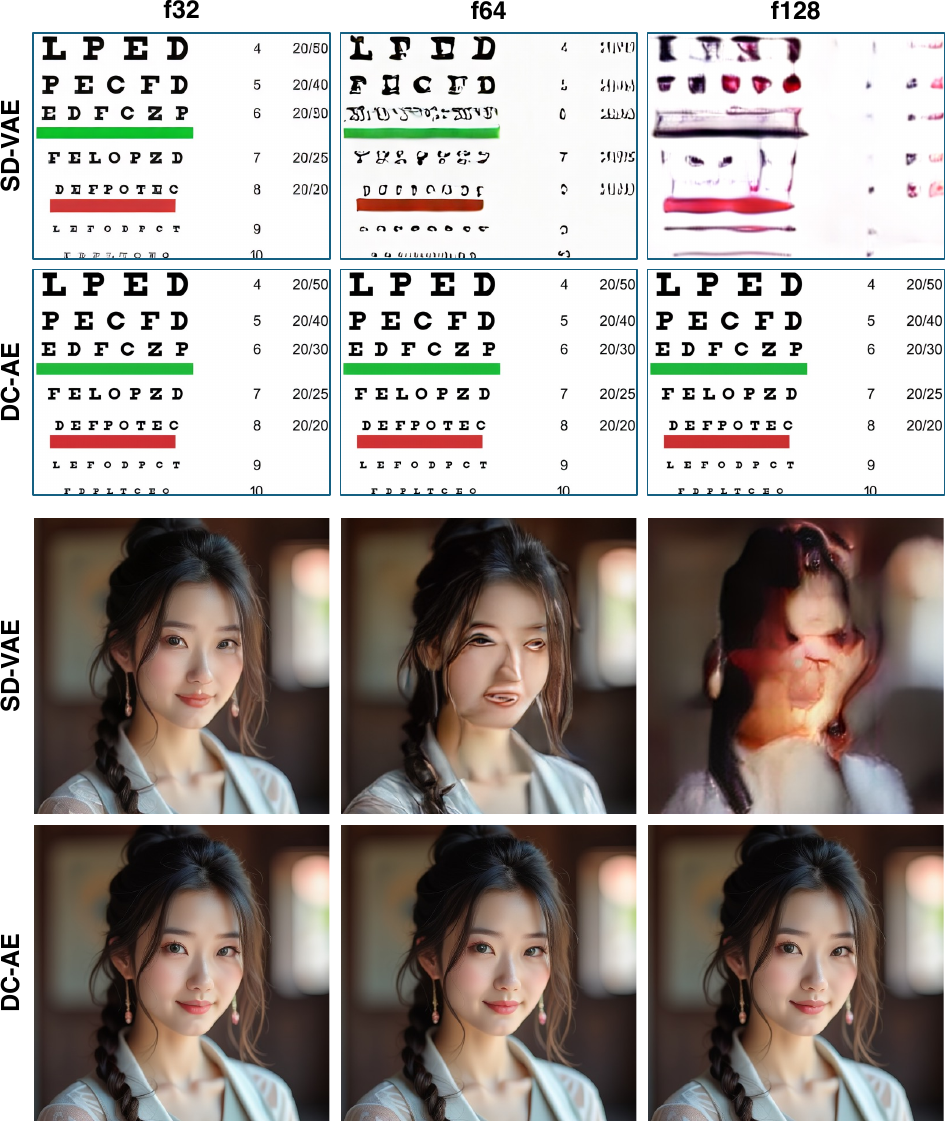}
    \vspace{-10pt}
    \caption{\textbf{Autoencoder Image Reconstruction Samples.}}
    \vspace{-10pt}
    \label{fig:ae_visualization}
\end{figure}

\begin{figure}[t]
    \centering
    \includegraphics[width=1\linewidth]{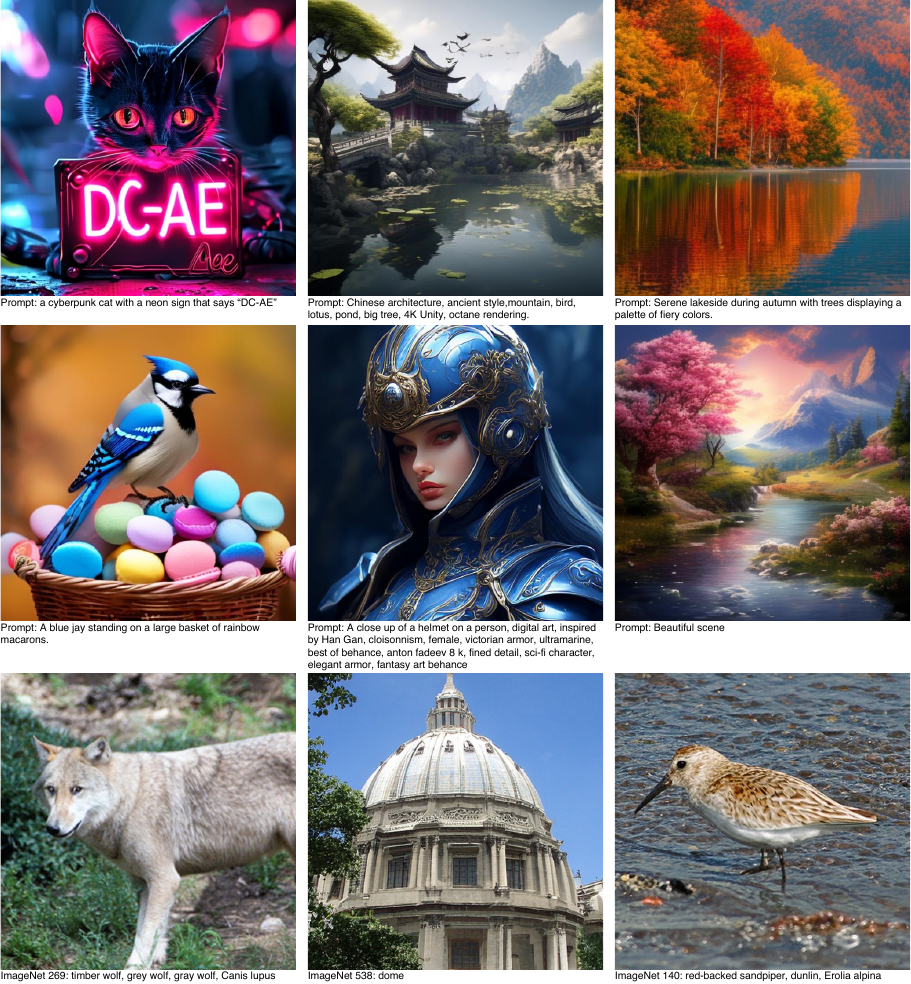}
    \vspace{-15pt}
    \caption{\textbf{Images Generated by Diffusion Model using Our \modelshort.}}
    \vspace{-10pt}
    \label{fig:diffusion_visualization}
\end{figure}

\paragraph{Efficiency Profiling.} We profile the training and inference throughput on the H100 GPU with PyTorch and TensorRT respectively. The latency is measured on the 3090 GPU with batch size 2. The training memory is profiled using PyTorch, assuming a batch size of 256. We use fp16 for all cases.

\subsection{Image Compression and Reconstruction}
Table~\ref{tab:ae_main} summarizes the results of \modelshort and SD-VAE \citep{rombach2022high} under various settings (f represents the spatial compression ratio and c denotes the number of latent channels). \modelshort provides significant reconstruction accuracy improvements than SD-VAE for all cases. For example, on ImageNet $512 \times 512$, \modelshort improves the rFID from 16.84 to 0.22 for the f64c128 autoencoder and 100.74 to 0.23 for the f128c512 autoencoder. 

\begin{wrapfigure}{r}{0.35\textwidth}
  \vspace{-30pt}
  \begin{center}
    \includegraphics[width=0.35\textwidth]{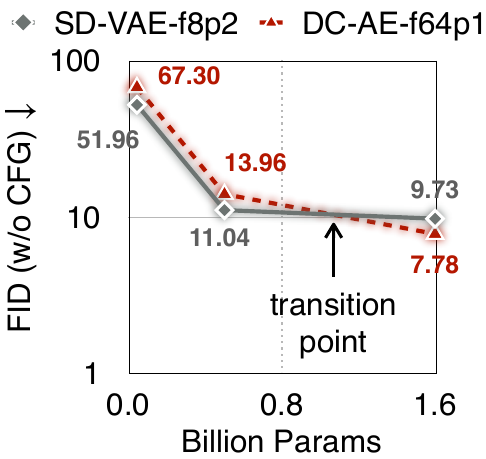}
  \end{center}
  \vspace{-10pt}
  \caption{\textbf{Model Scaling Results on ImageNet 512$\times$512 with UViT.} \modelshort-f64 benefits more from scaling up than SD-VAE-f8.}
  \vspace{-45pt}
  \label{fig:diffusion_scaling_up}
\end{wrapfigure}

In addition to the quantitative results, Figure~\ref{fig:ae_visualization} shows image reconstruction samples produced by SD-VAE and \modelshort. Reconstructed images by \modelshort demonstrate a better visual quality than SD-VAE's reconstructed images. In particular, for the f64 and f128 autoencoders,  \modelshort still maintains a good visual quality for small text and the human face. 

\subsection{Latent Diffusion Models}

We compare \modelshort with the widely used SD-VAE-f8 autoencoder \citep{rombach2022high} on various diffusion transformer models. For \modelshort, we always use a patch size of 1 (denoted as p1). For SD-VAE-f8, we follow the common setting and use a patch size of 2 or 4 (denoted as p2, p4). The results are summarized in Table~\ref{tab:diffusion_imagenet_main}, Table~\ref{tab:diffusion_t2i_main}, and Figure~\ref{fig:diffusion_scaling_up}. 

\vspace{-5pt}
\paragraph{ImageNet 512$\times$512.} As shown in Table~\ref{tab:diffusion_imagenet_main}, \modelshort-f32p1 consistently delivers better FID than SD-VAE-f8p2 on all diffusion transformer models. In addition, it has 4$\times$ fewer tokens than SD-VAE-f8p2, leading to 4.5$\times$ higher H100 training throughput and 4.8$\times$ higher H100 inference throughput for DiT-XL. We also observe that larger diffusion transformer models seem to benefit more from our \modelshort (Figure~\ref{fig:diffusion_scaling_up}). For example, \modelshort-f64p1 has a worse FID than SD-VAE-f8p2 on UViT-S but a better FID on UViT-2B. We conjecture it is because \modelshort-f64 has a larger latent channel number than SD-VAE-f8, thus needing more model capacity \citep{esser2024scaling}. 

Applying \modelshort to USiT models, we achieve highly competitive results compared with prior leading image generative models. For example, \modelshort-f32+USiT-2B achieves 1.72 FID on ImageNet 512$\times$512, outperforming the SOTA diffusion model EDM2-XXL and SOTA auto-regressive image generative models (MAGVIT-v2 and MAR-L). 

\vspace{-10pt}
\paragraph{Text-to-Image Generation.} Table~\ref{tab:diffusion_t2i_main} reports our text-to-image generation results. All models are trained for 100K iterations from scratch. Similar to prior cases, we observe \modelshort-f32p1 provides a better FID and a better CLIP Score than SD-VAE-f8p2. Figure~\ref{fig:diffusion_visualization} demonstrates samples generated by the diffusion models with our \modelshort, showing the capacity to synthesize high-quality images while being significantly more efficient than prior models.
\vspace{-5pt}
\section{Conclusion}\label{sec:conclusion}
\vspace{-5pt}
We accelerate high-resolution diffusion models by designing deep compression autoencoders to reduce the number of tokens. We proposed two techniques: \textit{residual autoencoding} and \textit{decoupled high-resolution adaptation} to address the challenges brought by the high compression ratio. The resulting new autoencoder family \modelshort demonstrated satisfactory reconstruction accuracy with a spatial compression ratio of up to 128. \modelshort also demonstrated significant training and inference efficiency improvements when applied to latent diffusion models. 

\section*{Acknowledgements}
We thank NVIDIA for donating the DGX machines. We thank MIT-IBM Watson AI Lab, MIT and Amazon Science Hub, MIT AI Hardware Program, and National Science Foundation for supporting this research.

\bibliography{main}

\begin{thebibliography}{58}
\providecommand{\natexlab}[1]{#1}
\providecommand{\url}[1]{\texttt{#1}}
\expandafter\ifx\csname urlstyle\endcsname\relax
  \providecommand{\doi}[1]{doi: #1}\else
  \providecommand{\doi}{doi: \begingroup \urlstyle{rm}\Url}\fi

\bibitem[Bao et~al.(2023)Bao, Nie, Xue, Cao, Li, Su, and Zhu]{bao2023all}
Fan Bao, Shen Nie, Kaiwen Xue, Yue Cao, Chongxuan Li, Hang Su, and Jun Zhu.
\newblock All are worth words: A vit backbone for diffusion models.
\newblock In \emph{Proceedings of the IEEE/CVF conference on computer vision and pattern recognition}, pp.\  22669--22679, 2023.

\bibitem[Cai et~al.(2020)Cai, Gan, Zhu, and Han]{cai2020tinytl}
Han Cai, Chuang Gan, Ligeng Zhu, and Song Han.
\newblock Tinytl: Reduce memory, not parameters for efficient on-device learning.
\newblock \emph{Advances in Neural Information Processing Systems}, 33:\penalty0 11285--11297, 2020.

\bibitem[Cai et~al.(2023)Cai, Li, Hu, Gan, and Han]{cai2023efficientvit}
Han Cai, Junyan Li, Muyan Hu, Chuang Gan, and Song Han.
\newblock Efficientvit: Lightweight multi-scale attention for high-resolution dense prediction.
\newblock In \emph{Proceedings of the IEEE/CVF International Conference on Computer Vision}, pp.\  17302--17313, 2023.

\bibitem[Cai et~al.(2024)Cai, Li, Zhang, Liu, and Han]{cai2024condition}
Han Cai, Muyang Li, Qinsheng Zhang, Ming-Yu Liu, and Song Han.
\newblock Condition-aware neural network for controlled image generation.
\newblock In \emph{Proceedings of the IEEE/CVF Conference on Computer Vision and Pattern Recognition}, pp.\  7194--7203, 2024.

\bibitem[Chen et~al.(2024{\natexlab{a}})Chen, Ge, Xie, Wu, Yao, Ren, Wang, Luo, Lu, and Li]{chen2024pixart}
Junsong Chen, Chongjian Ge, Enze Xie, Yue Wu, Lewei Yao, Xiaozhe Ren, Zhongdao Wang, Ping Luo, Huchuan Lu, and Zhenguo Li.
\newblock Pixart-$\sigma$: Weak-to-strong training of diffusion transformer for 4k text-to-image generation.
\newblock \emph{arXiv preprint arXiv:2403.04692}, 2024{\natexlab{a}}.

\bibitem[Chen et~al.(2024{\natexlab{b}})Chen, Jincheng, Chongjian, Yao, Xie, Wang, Kwok, Luo, Lu, and Li]{chenpixart}
Junsong Chen, YU~Jincheng, GE~Chongjian, Lewei Yao, Enze Xie, Zhongdao Wang, James Kwok, Ping Luo, Huchuan Lu, and Zhenguo Li.
\newblock Pixart-$\alpha$: Fast training of diffusion transformer for photorealistic text-to-image synthesis.
\newblock In \emph{International Conference on Learning Representations}, 2024{\natexlab{b}}.

\bibitem[Dai et~al.(2023)Dai, Hou, Ma, Tsai, Wang, Wang, Zhang, Vandenhende, Wang, Dubey, et~al.]{dai2023emu}
Xiaoliang Dai, Ji~Hou, Chih-Yao Ma, Sam Tsai, Jialiang Wang, Rui Wang, Peizhao Zhang, Simon Vandenhende, Xiaofang Wang, Abhimanyu Dubey, et~al.
\newblock Emu: Enhancing image generation models using photogenic needles in a haystack.
\newblock \emph{arXiv preprint arXiv:2309.15807}, 2023.

\bibitem[Deng et~al.(2009)Deng, Dong, Socher, Li, Li, and Fei-Fei]{deng2009imagenet}
Jia Deng, Wei Dong, Richard Socher, Li-Jia Li, Kai Li, and Li~Fei-Fei.
\newblock Imagenet: A large-scale hierarchical image database.
\newblock In \emph{2009 IEEE conference on computer vision and pattern recognition}, pp.\  248--255. Ieee, 2009.

\bibitem[Esser et~al.(2024)Esser, Kulal, Blattmann, Entezari, M{\"u}ller, Saini, Levi, Lorenz, Sauer, Boesel, et~al.]{esser2024scaling}
Patrick Esser, Sumith Kulal, Andreas Blattmann, Rahim Entezari, Jonas M{\"u}ller, Harry Saini, Yam Levi, Dominik Lorenz, Axel Sauer, Frederic Boesel, et~al.
\newblock Scaling rectified flow transformers for high-resolution image synthesis.
\newblock In \emph{Forty-first International Conference on Machine Learning}, 2024.

\bibitem[Fang et~al.(2024)Fang, Ma, and Wang]{fang2024structural}
Gongfan Fang, Xinyin Ma, and Xinchao Wang.
\newblock Structural pruning for diffusion models.
\newblock \emph{Advances in Neural Information Processing Systems}, 36, 2024.

\bibitem[He et~al.(2016)He, Zhang, Ren, and Sun]{he2016deep}
Kaiming He, Xiangyu Zhang, Shaoqing Ren, and Jian Sun.
\newblock Deep residual learning for image recognition.
\newblock In \emph{Proceedings of the IEEE conference on computer vision and pattern recognition}, pp.\  770--778, 2016.

\bibitem[He et~al.(2024)He, Liu, Liu, Wu, Zhou, and Zhuang]{he2024ptqd}
Yefei He, Luping Liu, Jing Liu, Weijia Wu, Hong Zhou, and Bohan Zhuang.
\newblock Ptqd: Accurate post-training quantization for diffusion models.
\newblock \emph{Advances in Neural Information Processing Systems}, 36, 2024.

\bibitem[Ho et~al.(2020)Ho, Jain, and Abbeel]{ho2020denoising}
Jonathan Ho, Ajay Jain, and Pieter Abbeel.
\newblock Denoising diffusion probabilistic models.
\newblock \emph{Advances in neural information processing systems}, 33:\penalty0 6840--6851, 2020.

\bibitem[Isola et~al.(2017)Isola, Zhu, Zhou, and Efros]{isola2017image}
Phillip Isola, Jun-Yan Zhu, Tinghui Zhou, and Alexei~A Efros.
\newblock Image-to-image translation with conditional adversarial networks.
\newblock In \emph{Proceedings of the IEEE conference on computer vision and pattern recognition}, pp.\  1125--1134, 2017.

\bibitem[Jayasumana et~al.(2024)Jayasumana, Ramalingam, Veit, Glasner, Chakrabarti, and Kumar]{jayasumana2024rethinking}
Sadeep Jayasumana, Srikumar Ramalingam, Andreas Veit, Daniel Glasner, Ayan Chakrabarti, and Sanjiv Kumar.
\newblock Rethinking fid: Towards a better evaluation metric for image generation.
\newblock In \emph{Proceedings of the IEEE/CVF Conference on Computer Vision and Pattern Recognition}, pp.\  9307--9315, 2024.

\bibitem[Karras et~al.(2019)Karras, Laine, and Aila]{karras2019style}
Tero Karras, Samuli Laine, and Timo Aila.
\newblock A style-based generator architecture for generative adversarial networks.
\newblock In \emph{Proceedings of the IEEE/CVF conference on computer vision and pattern recognition}, pp.\  4401--4410, 2019.

\bibitem[Karras et~al.(2024)Karras, Aittala, Lehtinen, Hellsten, Aila, and Laine]{karras2024analyzing}
Tero Karras, Miika Aittala, Jaakko Lehtinen, Janne Hellsten, Timo Aila, and Samuli Laine.
\newblock Analyzing and improving the training dynamics of diffusion models.
\newblock In \emph{Proceedings of the IEEE/CVF Conference on Computer Vision and Pattern Recognition}, pp.\  24174--24184, 2024.

\bibitem[Kirillov et~al.(2023)Kirillov, Mintun, Ravi, Mao, Rolland, Gustafson, Xiao, Whitehead, Berg, Lo, et~al.]{kirillov2023segment}
Alexander Kirillov, Eric Mintun, Nikhila Ravi, Hanzi Mao, Chloe Rolland, Laura Gustafson, Tete Xiao, Spencer Whitehead, Alexander~C Berg, Wan-Yen Lo, et~al.
\newblock Segment anything.
\newblock In \emph{Proceedings of the IEEE/CVF International Conference on Computer Vision}, pp.\  4015--4026, 2023.

\bibitem[Kynk{\"a}{\"a}nniemi et~al.(2019)Kynk{\"a}{\"a}nniemi, Karras, Laine, Lehtinen, and Aila]{kynkaanniemi2019improved}
Tuomas Kynk{\"a}{\"a}nniemi, Tero Karras, Samuli Laine, Jaakko Lehtinen, and Timo Aila.
\newblock Improved precision and recall metric for assessing generative models.
\newblock \emph{Advances in neural information processing systems}, 32, 2019.

\bibitem[Labs(2024)]{flux2024}
Black~Forest Labs.
\newblock Flux.
\newblock \emph{Online}, 2024.
\newblock URL \url{https://github.com/black-forest-labs/flux}.

\bibitem[Li et~al.(2024{\natexlab{a}})Li, Kamko, Akhgari, Sabet, Xu, and Doshi]{li2024playground}
Daiqing Li, Aleks Kamko, Ehsan Akhgari, Ali Sabet, Linmiao Xu, and Suhail Doshi.
\newblock Playground v2. 5: Three insights towards enhancing aesthetic quality in text-to-image generation.
\newblock \emph{arXiv preprint arXiv:2402.17245}, 2024{\natexlab{a}}.

\bibitem[Li et~al.(2022)Li, Lin, Meng, Ermon, Han, and Zhu]{li2022efficient}
Muyang Li, Ji~Lin, Chenlin Meng, Stefano Ermon, Song Han, and Jun-Yan Zhu.
\newblock Efficient spatially sparse inference for conditional gans and diffusion models.
\newblock \emph{Advances in neural information processing systems}, 35:\penalty0 28858--28873, 2022.

\bibitem[Li et~al.(2024{\natexlab{b}})Li, Cai, Cao, Zhang, Cai, Bai, Jia, Li, and Han]{li2024distrifusion}
Muyang Li, Tianle Cai, Jiaxin Cao, Qinsheng Zhang, Han Cai, Junjie Bai, Yangqing Jia, Kai Li, and Song Han.
\newblock Distrifusion: Distributed parallel inference for high-resolution diffusion models.
\newblock In \emph{Proceedings of the IEEE/CVF Conference on Computer Vision and Pattern Recognition}, pp.\  7183--7193, 2024{\natexlab{b}}.

\bibitem[Li et~al.(2024{\natexlab{c}})Li, Tian, Li, Deng, and He]{li2024autoregressive}
Tianhong Li, Yonglong Tian, He~Li, Mingyang Deng, and Kaiming He.
\newblock Autoregressive image generation without vector quantization.
\newblock \emph{arXiv preprint arXiv:2406.11838}, 2024{\natexlab{c}}.

\bibitem[Li et~al.(2023)Li, Liu, Lian, Yang, Dong, Kang, Zhang, and Keutzer]{li2023q}
Xiuyu Li, Yijiang Liu, Long Lian, Huanrui Yang, Zhen Dong, Daniel Kang, Shanghang Zhang, and Kurt Keutzer.
\newblock Q-diffusion: Quantizing diffusion models.
\newblock In \emph{Proceedings of the IEEE/CVF International Conference on Computer Vision}, pp.\  17535--17545, 2023.

\bibitem[Li et~al.(2024{\natexlab{d}})Li, Wang, Jin, Hu, Chemerys, Fu, Wang, Tulyakov, and Ren]{li2024snapfusion}
Yanyu Li, Huan Wang, Qing Jin, Ju~Hu, Pavlo Chemerys, Yun Fu, Yanzhi Wang, Sergey Tulyakov, and Jian Ren.
\newblock Snapfusion: Text-to-image diffusion model on mobile devices within two seconds.
\newblock \emph{Advances in Neural Information Processing Systems}, 36, 2024{\natexlab{d}}.

\bibitem[Liu et~al.(2024)Liu, Yu, Tan, and Wang]{liu2024linfusion}
Songhua Liu, Weihao Yu, Zhenxiong Tan, and Xinchao Wang.
\newblock Linfusion: 1 gpu, 1 minute, 16k image.
\newblock \emph{arXiv preprint arXiv:2409.02097}, 2024.

\bibitem[Liu et~al.(2023)Liu, Zhang, Ma, Peng, et~al.]{liu2023instaflow}
Xingchao Liu, Xiwen Zhang, Jianzhu Ma, Jian Peng, et~al.
\newblock Instaflow: One step is enough for high-quality diffusion-based text-to-image generation.
\newblock In \emph{The Twelfth International Conference on Learning Representations}, 2023.

\bibitem[Loshchilov(2017)]{loshchilov2017decoupled}
I~Loshchilov.
\newblock Decoupled weight decay regularization.
\newblock \emph{arXiv preprint arXiv:1711.05101}, 2017.

\bibitem[Lu et~al.(2022{\natexlab{a}})Lu, Zhou, Bao, Chen, Li, and Zhu]{lu2022dpm}
Cheng Lu, Yuhao Zhou, Fan Bao, Jianfei Chen, Chongxuan Li, and Jun Zhu.
\newblock Dpm-solver: A fast ode solver for diffusion probabilistic model sampling in around 10 steps.
\newblock \emph{Advances in Neural Information Processing Systems}, 35:\penalty0 5775--5787, 2022{\natexlab{a}}.

\bibitem[Lu et~al.(2022{\natexlab{b}})Lu, Zhou, Bao, Chen, Li, and Zhu]{lu2022dpm2}
Cheng Lu, Yuhao Zhou, Fan Bao, Jianfei Chen, Chongxuan Li, and Jun Zhu.
\newblock Dpm-solver++: Fast solver for guided sampling of diffusion probabilistic models.
\newblock \emph{arXiv preprint arXiv:2211.01095}, 2022{\natexlab{b}}.

\bibitem[Luo et~al.(2023)Luo, Tan, Huang, Li, and Zhao]{luo2023latent}
Simian Luo, Yiqin Tan, Longbo Huang, Jian Li, and Hang Zhao.
\newblock Latent consistency models: Synthesizing high-resolution images with few-step inference.
\newblock \emph{arXiv preprint arXiv:2310.04378}, 2023.

\bibitem[Ma et~al.(2024{\natexlab{a}})Ma, Goldstein, Albergo, Boffi, Vanden-Eijnden, and Xie]{ma2024sit}
Nanye Ma, Mark Goldstein, Michael~S Albergo, Nicholas~M Boffi, Eric Vanden-Eijnden, and Saining Xie.
\newblock Sit: Exploring flow and diffusion-based generative models with scalable interpolant transformers.
\newblock \emph{arXiv preprint arXiv:2401.08740}, 2024{\natexlab{a}}.

\bibitem[Ma et~al.(2024{\natexlab{b}})Ma, Fang, and Wang]{ma2024deepcache}
Xinyin Ma, Gongfan Fang, and Xinchao Wang.
\newblock Deepcache: Accelerating diffusion models for free.
\newblock In \emph{Proceedings of the IEEE/CVF Conference on Computer Vision and Pattern Recognition}, pp.\  15762--15772, 2024{\natexlab{b}}.

\bibitem[Martin et~al.(2017)Martin, Hubert, Thomas, Bernhard, and Sepp]{martin2017gans}
Heusel Martin, Ramsauer Hubert, Unterthiner Thomas, Nessler Bernhard, and Hochreiter Sepp.
\newblock Gans trained by a two time-scale update rule converge to a local nash equilibrium.
\newblock \emph{Advances in neural information processing systems}, 30:\penalty0 6626--6637, 2017.

\bibitem[Meng et~al.(2023)Meng, Rombach, Gao, Kingma, Ermon, Ho, and Salimans]{meng2023distillation}
Chenlin Meng, Robin Rombach, Ruiqi Gao, Diederik Kingma, Stefano Ermon, Jonathan Ho, and Tim Salimans.
\newblock On distillation of guided diffusion models.
\newblock In \emph{Proceedings of the IEEE/CVF Conference on Computer Vision and Pattern Recognition}, pp.\  14297--14306, 2023.

\bibitem[Neuhold et~al.(2017)Neuhold, Ollmann, Rota~Bulo, and Kontschieder]{neuhold2017mapillary}
Gerhard Neuhold, Tobias Ollmann, Samuel Rota~Bulo, and Peter Kontschieder.
\newblock The mapillary vistas dataset for semantic understanding of street scenes.
\newblock In \emph{Proceedings of the IEEE international conference on computer vision}, pp.\  4990--4999, 2017.

\bibitem[Peebles \& Xie(2023)Peebles and Xie]{peebles2023scalable}
William Peebles and Saining Xie.
\newblock Scalable diffusion models with transformers.
\newblock In \emph{Proceedings of the IEEE/CVF International Conference on Computer Vision}, pp.\  4195--4205, 2023.

\bibitem[Podell et~al.(2023)Podell, English, Lacey, Blattmann, Dockhorn, M{\"u}ller, Penna, and Rombach]{podell2023sdxl}
Dustin Podell, Zion English, Kyle Lacey, Andreas Blattmann, Tim Dockhorn, Jonas M{\"u}ller, Joe Penna, and Robin Rombach.
\newblock Sdxl: Improving latent diffusion models for high-resolution image synthesis.
\newblock \emph{arXiv preprint arXiv:2307.01952}, 2023.

\bibitem[Rombach et~al.(2022)Rombach, Blattmann, Lorenz, Esser, and Ommer]{rombach2022high}
Robin Rombach, Andreas Blattmann, Dominik Lorenz, Patrick Esser, and Bj{\"o}rn Ommer.
\newblock High-resolution image synthesis with latent diffusion models.
\newblock In \emph{Proceedings of the IEEE/CVF conference on computer vision and pattern recognition}, pp.\  10684--10695, 2022.

\bibitem[Salimans \& Ho(2022)Salimans and Ho]{salimans2022progressive}
Tim Salimans and Jonathan Ho.
\newblock Progressive distillation for fast sampling of diffusion models.
\newblock In \emph{International Conference on Learning Representations}, 2022.

\bibitem[Salimans et~al.(2016)Salimans, Goodfellow, Zaremba, Cheung, Radford, and Chen]{salimans2016improved}
Tim Salimans, Ian Goodfellow, Wojciech Zaremba, Vicki Cheung, Alec Radford, and Xi~Chen.
\newblock Improved techniques for training gans.
\newblock \emph{Advances in neural information processing systems}, 29, 2016.

\bibitem[Shi et~al.(2016)Shi, Caballero, Husz{\'a}r, Totz, Aitken, Bishop, Rueckert, and Wang]{shi2016real}
Wenzhe Shi, Jose Caballero, Ferenc Husz{\'a}r, Johannes Totz, Andrew~P Aitken, Rob Bishop, Daniel Rueckert, and Zehan Wang.
\newblock Real-time single image and video super-resolution using an efficient sub-pixel convolutional neural network.
\newblock In \emph{Proceedings of the IEEE conference on computer vision and pattern recognition}, pp.\  1874--1883, 2016.

\bibitem[Shih et~al.(2024)Shih, Belkhale, Ermon, Sadigh, and Anari]{shih2024parallel}
Andy Shih, Suneel Belkhale, Stefano Ermon, Dorsa Sadigh, and Nima Anari.
\newblock Parallel sampling of diffusion models.
\newblock \emph{Advances in Neural Information Processing Systems}, 36, 2024.

\bibitem[Song et~al.(2021)Song, Meng, and Ermon]{songdenoising}
Jiaming Song, Chenlin Meng, and Stefano Ermon.
\newblock Denoising diffusion implicit models.
\newblock In \emph{International Conference on Learning Representations}, 2021.

\bibitem[Song et~al.(2023)Song, Dhariwal, Chen, and Sutskever]{song2023consistency}
Yang Song, Prafulla Dhariwal, Mark Chen, and Ilya Sutskever.
\newblock Consistency models.
\newblock In \emph{International Conference on Machine Learning}, pp.\  32211--32252. PMLR, 2023.

\bibitem[Tang et~al.(2024)Tang, Tang, Luo, Wang, and Chang]{tang2024accelerating}
Zhiwei Tang, Jiasheng Tang, Hao Luo, Fan Wang, and Tsung-Hui Chang.
\newblock Accelerating parallel sampling of diffusion models.
\newblock In \emph{Forty-first International Conference on Machine Learning}, 2024.

\bibitem[Wang et~al.(2024)Wang, Fang, Li, and Yang]{wang2024pipefusion}
Jiannan Wang, Jiarui Fang, Aoyu Li, and PengCheng Yang.
\newblock Pipefusion: Displaced patch pipeline parallelism for inference of diffusion transformer models.
\newblock \emph{arXiv preprint arXiv:2405.14430}, 2024.

\bibitem[Yin et~al.(2024{\natexlab{a}})Yin, Gharbi, Park, Zhang, Shechtman, Durand, and Freeman]{yin2024improved}
Tianwei Yin, Micha{\"e}l Gharbi, Taesung Park, Richard Zhang, Eli Shechtman, Fredo Durand, and William~T Freeman.
\newblock Improved distribution matching distillation for fast image synthesis.
\newblock \emph{arXiv preprint arXiv:2405.14867}, 2024{\natexlab{a}}.

\bibitem[Yin et~al.(2024{\natexlab{b}})Yin, Gharbi, Zhang, Shechtman, Durand, Freeman, and Park]{yin2024one}
Tianwei Yin, Micha{\"e}l Gharbi, Richard Zhang, Eli Shechtman, Fredo Durand, William~T Freeman, and Taesung Park.
\newblock One-step diffusion with distribution matching distillation.
\newblock In \emph{Proceedings of the IEEE/CVF Conference on Computer Vision and Pattern Recognition}, pp.\  6613--6623, 2024{\natexlab{b}}.

\bibitem[Yu et~al.(2023)Yu, Lezama, Gundavarapu, Versari, Sohn, Minnen, Cheng, Birodkar, Gupta, Gu, et~al.]{yu2023language}
Lijun Yu, Jos{\'e} Lezama, Nitesh~B Gundavarapu, Luca Versari, Kihyuk Sohn, David Minnen, Yong Cheng, Vighnesh Birodkar, Agrim Gupta, Xiuye Gu, et~al.
\newblock Language model beats diffusion--tokenizer is key to visual generation.
\newblock \emph{arXiv preprint arXiv:2310.05737}, 2023.

\bibitem[Zhang \& Chen(2023)Zhang and Chen]{zhangfast}
Qinsheng Zhang and Yongxin Chen.
\newblock Fast sampling of diffusion models with exponential integrator.
\newblock In \emph{The Eleventh International Conference on Learning Representations}, 2023.

\bibitem[Zhang et~al.(2023)Zhang, Tao, and Chen]{zhanggddim}
Qinsheng Zhang, Molei Tao, and Yongxin Chen.
\newblock gddim: Generalized denoising diffusion implicit models.
\newblock In \emph{International Conference on Learning Representations}, 2023.

\bibitem[Zhang et~al.(2018)Zhang, Isola, Efros, Shechtman, and Wang]{zhang2018unreasonable}
Richard Zhang, Phillip Isola, Alexei~A Efros, Eli Shechtman, and Oliver Wang.
\newblock The unreasonable effectiveness of deep features as a perceptual metric.
\newblock In \emph{Proceedings of the IEEE conference on computer vision and pattern recognition}, pp.\  586--595, 2018.

\bibitem[Zhao et~al.(2024{\natexlab{a}})Zhao, Fang, Liu, Rui, Soedarmadji, Li, Lin, Dai, Yan, Yang, et~al.]{zhao2024vidit}
Tianchen Zhao, Tongcheng Fang, Enshu Liu, Wan Rui, Widyadewi Soedarmadji, Shiyao Li, Zinan Lin, Guohao Dai, Shengen Yan, Huazhong Yang, et~al.
\newblock Vidit-q: Efficient and accurate quantization of diffusion transformers for image and video generation.
\newblock \emph{arXiv preprint arXiv:2406.02540}, 2024{\natexlab{a}}.

\bibitem[Zhao et~al.(2024{\natexlab{b}})Zhao, Bai, Rao, Zhou, and Lu]{zhao2024unipc}
Wenliang Zhao, Lujia Bai, Yongming Rao, Jie Zhou, and Jiwen Lu.
\newblock Unipc: A unified predictor-corrector framework for fast sampling of diffusion models.
\newblock \emph{Advances in Neural Information Processing Systems}, 36, 2024{\natexlab{b}}.

\bibitem[Zheng et~al.(2023)Zheng, Lu, Chen, and Zhu]{zheng2023dpm}
Kaiwen Zheng, Cheng Lu, Jianfei Chen, and Jun Zhu.
\newblock Dpm-solver-v3: Improved diffusion ode solver with empirical model statistics.
\newblock \emph{Advances in Neural Information Processing Systems}, 36:\penalty0 55502--55542, 2023.

\bibitem[Zhu et~al.(2023)Zhu, Feng, Chen, Bao, Wang, Chen, Yuan, and Hua]{zhu2023designing}
Zixin Zhu, Xuelu Feng, Dongdong Chen, Jianmin Bao, Le~Wang, Yinpeng Chen, Lu~Yuan, and Gang Hua.
\newblock Designing a better asymmetric vqgan for stablediffusion.
\newblock \emph{arXiv preprint arXiv:2306.04632}, 2023.

\end{thebibliography}
\bibliographystyle{iclr2025_conference}

\newpage

\appendix
\appendix

\section{\modelshort Architecture and Training Details}

\begin{figure}[ht]
    \centering
    \includegraphics[width=0.8\linewidth]{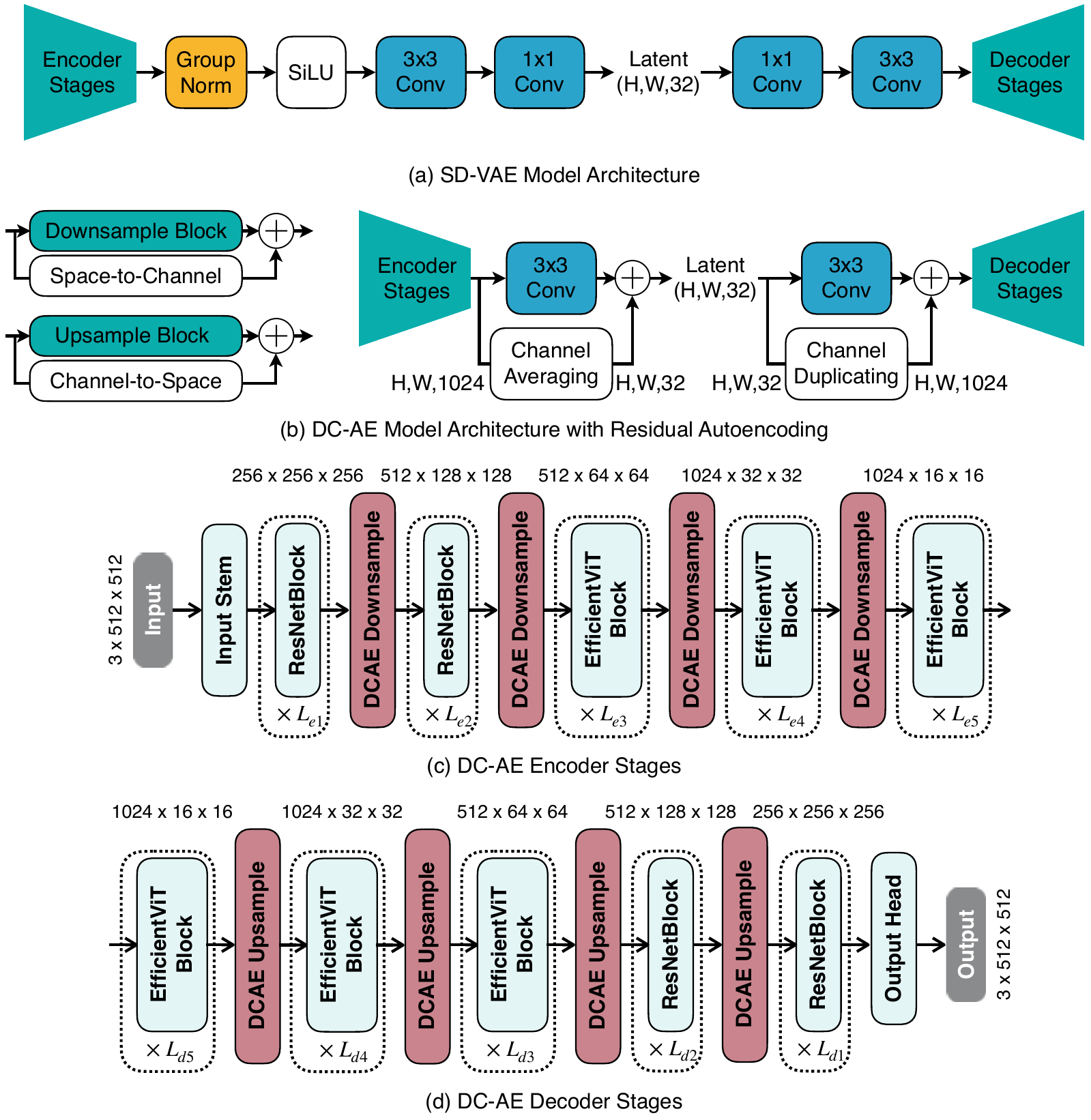}
    \vspace{-5pt}
    \caption{\textbf{Detailed Architecture of SD-VAE, DC-AE, DC-AE Encoder, and DC-AE Decoder Stages.} }
    \vspace{-5pt}
    \label{fig:dc_ae_detailed_arch}
\end{figure}

We present the detailed architecture of SD-VAE, \modelshort, \modelshort encoder, and \modelshort decoder stages in Figure \ref{fig:dc_ae_detailed_arch} to complement Figure \ref{fig:method_arch}.

We use the AdamW optimizer \citep{loshchilov2017decoupled} for all training phases.

In phase 1 (low-resolution full training), we use a constant learning rate of 6.4e-5 with a weight decay of 0.1, and AdamW betas of (0.9, 0.999). We use L1 loss and LPIPS loss \citep{zhang2018unreasonable}.

In phase 2 (high-resolution latent adaptation), we use a constant learning rate of 1.6e-5, a weight decay of 0.001, and AdamW betas of (0.9, 0.999). We use the same loss as phase 1. 

In phase 3 (low-resolution local refinement), we use a constant learning rate of 5.4e-5, and AdamW betas of (0.5, 0.9). We use L1 loss, LPIPS loss \citep{zhang2018unreasonable}, and PatchGAN loss \citep{isola2017image}.

\section{Ablation Study on Training Different Numbers of Layers}

\begin{figure}[ht]
    \centering
    \includegraphics[width=0.9\linewidth]{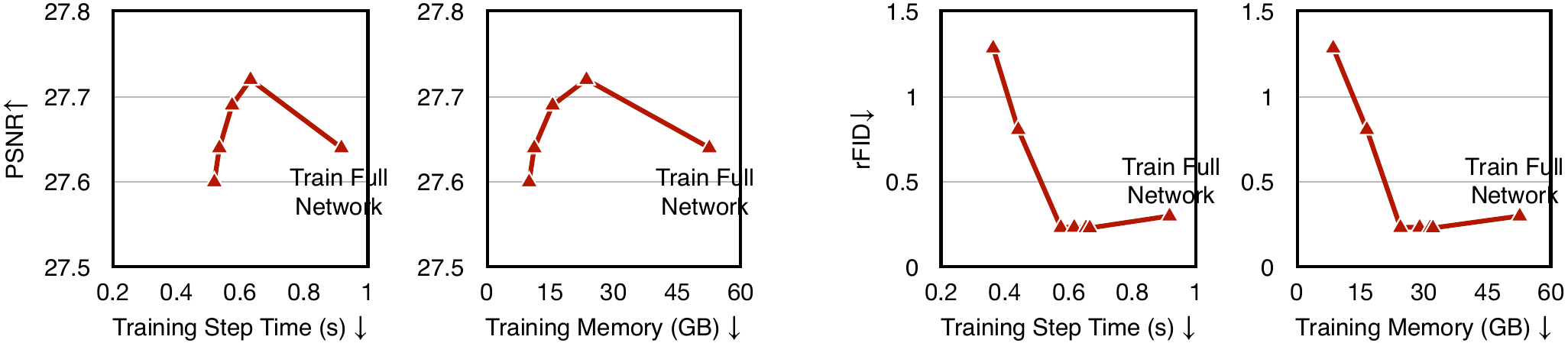}
    \vspace{-5pt}
    \caption{\textbf{Ablation Study on Training Different Numbers of Layers in Phase 2 (Left) and Phase 3 (Right).}}
    \vspace{-5pt}
    \label{fig:ablation_phase2_phase3}
\end{figure}

Figure \ref{fig:ablation_phase2_phase3} presents the ablation study on training different numbers of layers in phase 2 (high-resolution latent adaptation) and phase 3 (low-resolution local refinement).

\section{Additional Image Reconstruction Results}

Table~\ref{tab:ae_low_compression} reports the reconstruction results under the low spatial-compression ratio setting. \modelshort delivers slightly better results than SD-VAE under this setting. 

\begin{table}[ht]
\small\centering\setlength{\tabcolsep}{3pt}
\begin{tabular}{l | c | g | g g g g }
\toprule
\rowcolor{white} \textbf{ImageNet 256$\times$256} & Latent Shape & Autoencoder & rFID $\downarrow$ & PSNR $\uparrow$ & SSIM $\uparrow$ & LPIPS $\downarrow$ \\
 \midrule
\rowcolor{white} \multirow{2}{*}{f8c4} & \multirow{2}{*}{32$\times$32$\times$4}
 & SD-VAE \tablecite{rombach2022high} & 0.63 & 24.99 & 0.71 & 0.063 \\
 & & \modelshort                        & \textbf{0.46} & \textbf{25.46} & \textbf{0.73} & \textbf{0.057} \\
\bottomrule
\end{tabular}
\vspace{-5pt}
\caption{\textbf{Image Reconstruction Results under the Low Spatial-Compression Ratio Setting.}}
\vspace{-5pt}
\label{tab:ae_low_compression}
\end{table}

\section{Latent Scaling and Shifting Factors}

Following the common practice \citep{rombach2022high,peebles2023scalable,bao2023all,esser2024scaling,flux2024,chenpixart,chen2024pixart}, we normalize the latent space of our autoencoders to apply to latent diffusion models. Given a dataset, we compute the root mean square of the latent features and use its multiplicative inverse as the scaling factor for our autoencoders. We do not use the shifting factor for our autoencoders. 

\section{Diffusion Model Architecture Details}
In addition to existing UViT models, we scaled the model up to 1.6B parameters, with a depth of 28, a hidden dimension of 2048, and 32 heads. We denote this model as UViT-2B.

\section{Diffusion Sampling Hyperparameters}

For the DiT models, we use the DDPM \citep{ho2020denoising} sampler from the DiT \citep{peebles2023scalable} codebase with 250 sampling steps and a guidance scale of 1.3.

For the UViT models, we use the DPMSolver \citep{lu2022dpm} sampler with 30 sampling steps and a guidance scale of 1.5.

\begin{figure}[ht]
    \centering
    \includegraphics[width=1\linewidth]{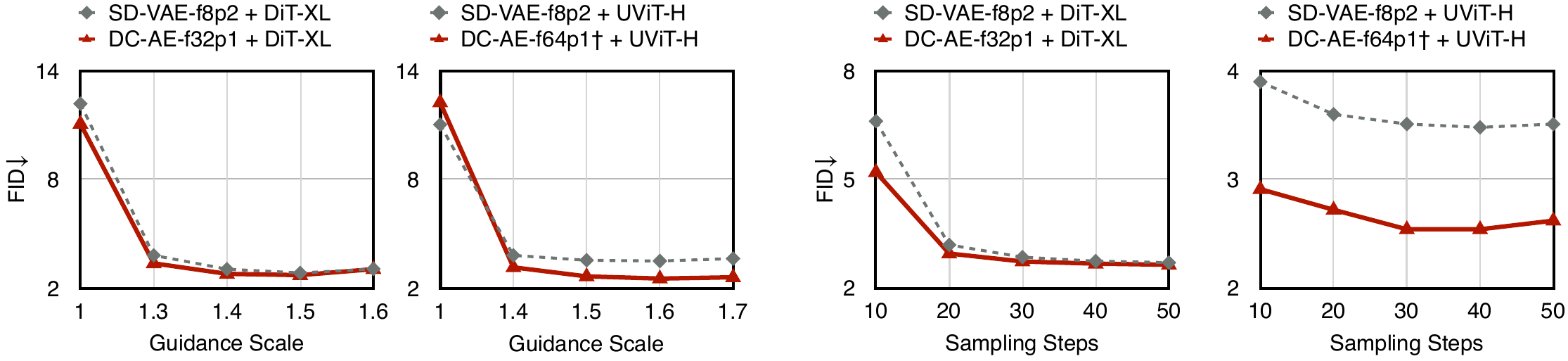}
    \caption{\textbf{Ablation Study on Diffusion Sampling Hyperparameters.} We use the DPMSolver sampler for both DiT-XL and UViT-H. \modelshort provides significant speedup over the baseline models while maintaining the generation performance under different diffusion sampling hyperparameters. }
    \label{fig:diffusion_sampling_params}
\end{figure}

\section{High-Resolution Image Generation Results}

\begin{table}[t]
\small\centering\setlength{\tabcolsep}{1pt}
\resizebox{1\linewidth}{!}{
\begin{tabular}{l | g | g | g | g g | g | g | g | g g}
\toprule
\multicolumn{10}{l}{\textbf{FFHQ 1024$\times$1024 (Unconditional) \& MJHQ 1024$\times$1024 (Class-Conditional)}} \\
\midrule
\rowcolor{white} Diffusion & & Patch & & \multicolumn{2}{c|}{Throughput (image/s) $\uparrow$} & Latency & Memory & FFHQ FID $\downarrow$ & \multicolumn{2}{c}{MJHQ FID $\downarrow$} \\
\rowcolor{white} Model & \multirow{-2}{*}{Autoencoder} & Size & \multirow{-2}{*}{NFE} & Training & Inference & (ms) $\downarrow$ & (GB) $\downarrow$ & w/o CFG & w/o CFG & w/ CFG \\
\midrule
\rowcolor{white} & SD3-VAE-f8 \tablecite{esser2024scaling}                 & 2 & 250 &   83 &  1.63 & 3554 & 41.4 & 46.28 & 109.43 & 103.02 \\
\rowcolor{white} & Flux-VAE-f8 \tablecite{flux2024}                        & 2 & 250 &   83 &  1.63 & 3554 & 41.4 & 59.15 & 143.16 & 139.06 \\
\cmidrule{2-11}
\rowcolor{white} & SDXL-VAE-f8 \tablecite{podell2023sdxl}                  & 2 & 250 &   84 &  1.67 & 3530 & 41.2 & 16.82 & 49.00 & 39.21 \\
\rowcolor{white} & Asym-VAE-f8 \tablecite{zhu2023designing}                & 2 & 250 &   84 &  1.67 & 3530 & 41.2 & 17.10 & 48.30 & 38.35 \\
\cmidrule{2-11}
\rowcolor{white} &                                                         & 2 & 250 &   84 &  1.67 & 3530 & 41.2 & 16.98 & 48.05 & 38.19 \\
\rowcolor{white} & \multirow{-2}{*}{SD-VAE-f8 \tablecite{rombach2022high}} & 4 & 250 &  470 & 11.13 &  632 & 10.7 & 23.81 & 60.94 & 51.29 \\ 
\cmidrule{2-11}
& \modelshort-f32                                                          & 1 & 250 &  475 & 11.15 &  634 & 10.7 & 13.65 & 34.35 & 27.20 \\
& \;\;\modelshort-f32$^\ddag$                                              & 1 & 250 &  475 & 11.15 &  634 & 10.7 & \textbf{11.39} & \textbf{28.36} & \textbf{21.89} \\
\multirow{-9}{*}{DiT-S \tablecite{peebles2023scalable}}
& \modelshort-f64                                                          & 1 & 250 & \textbf{2085} & \textbf{50.26} & \textbf{230} & \textbf{3.1} & 26.88 & 61.30 & 53.38 \\
\bottomrule
\toprule
\multicolumn{10}{l}{\textbf{MapillaryVistas 
 2048$\times$2048 (Unconditional)}} \\
\midrule
\rowcolor{white} Diffusion & & Patch & & \multicolumn{2}{c|}{Throughput (image/s) $\uparrow$} & Latency & Memory & \multicolumn{3}{c}{MapillaryVistas FID $\downarrow$} \\
\rowcolor{white} Model & \multirow{-2}{*}{Autoencoder} & Size & \multirow{-2}{*}{NFE} & Training & Inference & (ms) $\downarrow$ & (GB) $\downarrow$ & \multicolumn{3}{c}{w/o CFG}\\
\midrule
\rowcolor{white} & SD-VAE-f8 \tablecite{rombach2022high}                  & 4 & 250 &    84 &  1.64 & 3561 & 41.4 & \multicolumn{3}{c}{69.50} \\
\cmidrule{2-11}
\multirow{-2}{*}{DiT-S \tablecite{peebles2023scalable}} & \modelshort-f64 & 1 & 250 & \textbf{459} & \textbf{10.91} & \textbf{639} & \textbf{11.0} & \multicolumn{3}{g}{\textbf{59.55}} \\ 
\bottomrule
\end{tabular}
}
\vspace{-5pt}
\caption{\textbf{1024$\times$1024 and 2048$\times$2048 Image Generation Results.} $^\ddag$ represents the model is trained with 4$\times$ batch size (i.e., 256 $\rightarrow$ 1024).}
\label{tab:diffusion_hr_main}
\end{table}

Apart from ImageNet 512$\times$512, we also test our models for higher-resolution image generation. As shown in Table~\ref{tab:diffusion_hr_main}, we have a similar finding where \modelshort-f32p1 achieves better FID than SD-VAE-f8p2 for all cases. 

\section{Image Generation Results with Other Evaluation Metrics}

\begin{table}[t]
\small\centering\setlength{\tabcolsep}{2pt}
\resizebox{1\linewidth}{!}{
\begin{tabular}{l | g | g | g | g | g g | g g | g g | g g | g g}
\toprule
\rowcolor{white} Diffusion &                               & Patch & & Inference  & \multicolumn{2}{c|}{FID $\downarrow$} & \multicolumn{2}{c|}{Inception Score $\uparrow$} & \multicolumn{2}{c|}{Precision $\uparrow$} & \multicolumn{2}{c|}{Recall $\uparrow$} & \multicolumn{2}{c}{CMMD $\downarrow$} \\
\rowcolor{white} Model     & \multirow{-2}{*}{Autoencoder} & Size  & \multirow{-2}{*}{NFE} & Throughput & w/o CFG & w/ CFG                      & w/o CFG & w/ CFG                                & w/o CFG & w/ CFG                          & w/o CFG & w/ CFG                       & w/o CFG & w/ CFG \\
\midrule
\rowcolor{white} & SD3-VAE-f8 \tablecite{esser2024scaling}                      & 2 &  30 &   49.73 & 164.34 & 143.82 &   6.07 &   7.53 & 0.06 & 0.09 & 0.31 & 0.39 & 3.13 & 2.94 \\
\rowcolor{white} & Flux-VAE-f8 \tablecite{flux2024}                             & 2 &  30 &   49.73 & 106.07 &  84.73 &  13.39 &  17.71 & 0.28 & 0.37 & 0.39 & 0.42 & 1.90 & 1.67 \\
\cmidrule{2-15}
\rowcolor{white} & SDXL-VAE-f8 \tablecite{podell2023sdxl}                       & 2 &  30 &   49.85 &  51.03 &  26.38 &  27.58 &  56.72 & 0.57 & 0.74 & 0.58 & 0.50 & 1.35 & 1.05 \\
\rowcolor{white} & Asym-VAE-f8 \tablecite{zhu2023designing}                     & 2 &  30 &   49.85 &  52.68 &  25.14 &  30.22 &  65.27 & 0.58 & 0.74 & 0.62 & 0.51 & 1.09 & 0.80 \\
\rowcolor{white} & SD-VAE-f8 \tablecite{rombach2022high}                        & 2 &  30 &   49.85 &  51.96 &  24.57 &  30.37 &  65.73 & 0.57 & 0.74 & 0.64 & 0.52 & 1.23 & 0.91 \\
\rowcolor{white} & SD-VAE-f16 \tablecite{rombach2022high}                       & 2 &  30 &  214.68 &  76.86 &  44.22 &  21.38 &  43.35 & 0.43 & 0.62 & 0.60 & 0.55 & 1.83 & 1.46 \\
\rowcolor{white} & SD-VAE-f32 \tablecite{rombach2022high}                       & 1 &  30 &  214.72 &  70.23 &  38.63 &  23.07 &  47.72 & 0.46 & 0.64 & 0.58 & 0.56 & 1.71 & 1.36 \\
\cmidrule{2-15}
& \modelshort-f32                                                               & 1 &  30 &  214.17 &  46.12 &  18.08 &  34.82 &  84.73 & 0.59 & 0.76 & 0.66 & 0.56 & 1.00 & 0.70 \\
& \modelshort-f64                                                               & 1 &  30 &  896.23 &  67.30 &  35.96 &  24.55 &  52.86 & 0.44 & 0.64 & 0.60 & 0.56 & 1.44 & 1.14 \\
\multirow{-11}{*}{UViT-S \tablecite{bao2023all}} & \:\:\modelshort-f64$^\dag$   & 1 &  30 &  896.23 &  61.84 &  30.63 &  27.28 &  61.76 & 0.47 & 0.67 & 0.63 & 0.56 & 1.35 & 1.04 \\
\midrule
\midrule
\rowcolor{white} & Flux-VAE-f8 \tablecite{flux2024}                             & 2 & 250 &    0.83 &  27.35 &   8.72 &  53.09 & 130.20 & 0.68 & 0.83 & 0.61 & 0.48 & 0.54 & 0.30 \\
\cmidrule{2-15}
\rowcolor{white} & Asym-VAE-f8 \tablecite{zhu2023designing}                     & 2 & 250 &    0.85 &  11.39 &   2.97 & 108.70 & 241.10 & 0.75 & 0.83 & 0.65 & 0.53 & 0.37 & 0.20 \\
\rowcolor{white} & SD-VAE-f8 \tablecite{rombach2022high}                        & 2 & 250 &    0.85 &  12.03 &   3.04 & 105.25 & 240.82 & 0.75 & 0.84 & 0.64 & 0.54 & 0.43 & 0.25 \\
\cmidrule{2-15}
& \modelshort-f32                                                               & 1 & 250 &    4.03 &   9.56 &   2.84 & 117.49 & 226.98 & 0.75 & 0.82 & 0.64 & 0.55 & 0.34 & 0.22 \\
\multirow{-5}{*}{DiT-XL \tablecite{peebles2023scalable}}
& \:\:\modelshort-f32$^\ddag$                                                   & 1 & 250 &    4.03 &   6.88 &   2.41 & 141.07 & 263.56 & 0.76 & 0.82 & 0.63 & 0.56 & 0.29 & 0.18 \\
\midrule
\midrule
\rowcolor{white} & Flux-VAE-f8 \tablecite{flux2024}                             & 2 &  30 &    5.82 &  30.91 &  12.63 &  56.72 & 127.93 & 0.64 & 0.76 & 0.59 & 0.49 & 0.50 & 0.31 \\
\cmidrule{2-15}
\rowcolor{white} & Asym-VAE-f8 \tablecite{zhu2023designing}                     & 2 &  30 &    5.85 &  11.36 &   3.51 & 124.24 & 249.21 & 0.75 & 0.82 & 0.61 & 0.53 & 0.32 & 0.20 \\
\rowcolor{white} & SD-VAE-f8 \tablecite{rombach2022high}                        & 2 &  30 &    5.85 &  11.04 &   3.55 & 125.08 & 250.66 & 0.75 & 0.82 & 0.61 & 0.53 & 0.39 & 0.26 \\
\cmidrule{2-15}
& \modelshort-f32                                                               & 1 &  30 &   27.03 &   9.83 &   2.53 & 121.91 & 255.07 & 0.76 & 0.83 & 0.65 & 0.54 & 0.34 & 0.20 \\
& \modelshort-f64                                                               & 1 &  30 &  111.77 &  13.96 &   3.01 &  99.20 & 229.16 & 0.73 & 0.83 & 0.64 & 0.53 & 0.50 & 0.31 \\
\multirow{-7}{*}{UViT-H \tablecite{bao2023all}} & \:\:\modelshort-f64$^\dag$    & 1 &  30 &  111.77 &  12.26 &   2.66 & 109.20 & 239.82 & 0.73 & 0.82 & 0.67 & 0.57 & 0.43 & 0.27 \\
\midrule
\midrule
\rowcolor{white} & Flux-VAE-f8 \tablecite{flux2024}                             & 2 &  30 &    2.58 &  25.03 &  10.12 &  74.04 & 161.29 & 0.67 & 0.78 & 0.58 & 0.51 & 0.38 & 0.24 \\
\cmidrule{2-15}
\rowcolor{white} & Asym-VAE-f8 \tablecite{zhu2023designing}                     & 2 &  30 &    2.62 &   9.87 &   3.62 & 131.95 & 258.63 & 0.76 & 0.83 & 0.59 & 0.52 & 0.30 & 0.19 \\
\rowcolor{white} & SD-VAE-f8 \tablecite{rombach2022high}                        & 2 &  30 &    2.62 &   9.73 &   3.57 & 132.86 & 260.50 & 0.76 & 0.83 & 0.59 & 0.52 & 0.37 & 0.24 \\
\cmidrule{2-15}
& \modelshort-f32                                                               & 1 &  30 &   11.08 &   8.13 &   2.30 & 135.44 & 272.73 & 0.76 & 0.82 & 0.66 & 0.56 & 0.30 & 0.17 \\
& \modelshort-f64                                                               & 1 &  30 &   45.55 &   7.78 &   2.47 & 138.11 & 280.49 & 0.77 & 0.84 & 0.63 & 0.54 & 0.35 & 0.22 \\
\multirow{-7}{*}{UViT-2B \tablecite{bao2023all}} & \:\:\modelshort-f64$^\dag$   & 1 &  30 &   45.55 &   6.50 &   2.25 & 152.35 & 293.45 & 0.77 & 0.83 & 0.65 & 0.56 & 0.31 & 0.19 \\
\midrule
\midrule
\rowcolor{white} MAGVIT-v2 \tablecite{yu2023language} & -                       & - &   - &       - &   3.07 &   1.91 & 213.1  & 324.3  &    - &    - &    - &    - &    - &    - \\
\rowcolor{white} EDM2-XXL \tablecite{karras2024analyzing} & -                   & - &   - &       - &   1.91 &   1.81 &      - &      - &    - &    - &    - &    - &    - &    - \\
\rowcolor{white} MAR-L \tablecite{li2024autoregressive} & -                     & - &   - &       - &   2.74 &   1.73 & 205.2  & 279.9  &    - &    - &    - &    - &    - &    - \\
\midrule
SiT-XL \tablecite{ma2024sit} & \modelshort-f32                                  & 1 &   - &       - &   7.47 &   2.41 & 131.37 & 237.71 & 0.77 & 0.82 & 0.65 & 0.58 & 0.36 & 0.23 \\
USiT-H & \modelshort-f32                                                        & 1 &   - &       - &   3.80 &   1.89 & 174.58 & 252.35 & 0.78 & 0.82 & 0.64 & 0.60 & 0.24 & 0.18 \\
USiT-2B & \modelshort-f32                                                       & 1 &   - &       - &   2.90 &   1.72 & 187.68 & 248.10 & 0.79 & 0.82 & 0.63 & 0.61 & 0.21 & 0.17 \\
\bottomrule
\end{tabular}
}
\vspace{-5pt}
\caption{\textbf{Class-Conditional Image Generation Results on ImageNet 512$\times$512 with More Evaluation Metrics.} $^\dag$ represents the model is trained for 4$\times$ training iterations (i.e., 500K $\rightarrow$ 2,000K iterations). $^\ddag$ represents the model is trained with 4$\times$ batch size (i.e., 256 $\rightarrow$ 1024). `NFE' denotes the number of functional evaluations. The NFEs for SiT \citep{ma2024sit} and USiT models are left blank as they use an adaptive-step evaluation scheduler.}
\label{tab:diffusion_imagenet_complete}
\end{table}

Table \ref{tab:diffusion_imagenet_complete} presents a comprehensive evaluation of different diffusion models and autoencoders on ImageNet 512$\times$512. The evaluation metrics include FID \citep{martin2017gans}, inception score (IS) \citep{salimans2016improved}, precision, recall \citep{kynkaanniemi2019improved}, and CMMD \citep{jayasumana2024rethinking}. Our \modelshort consistently delivers significant efficiency improvements while maintaining the generation performance under different evaluation metrics.

\section{Additional Samples}

\begin{figure}[ht]
    \centering
    \includegraphics[width=1\linewidth]{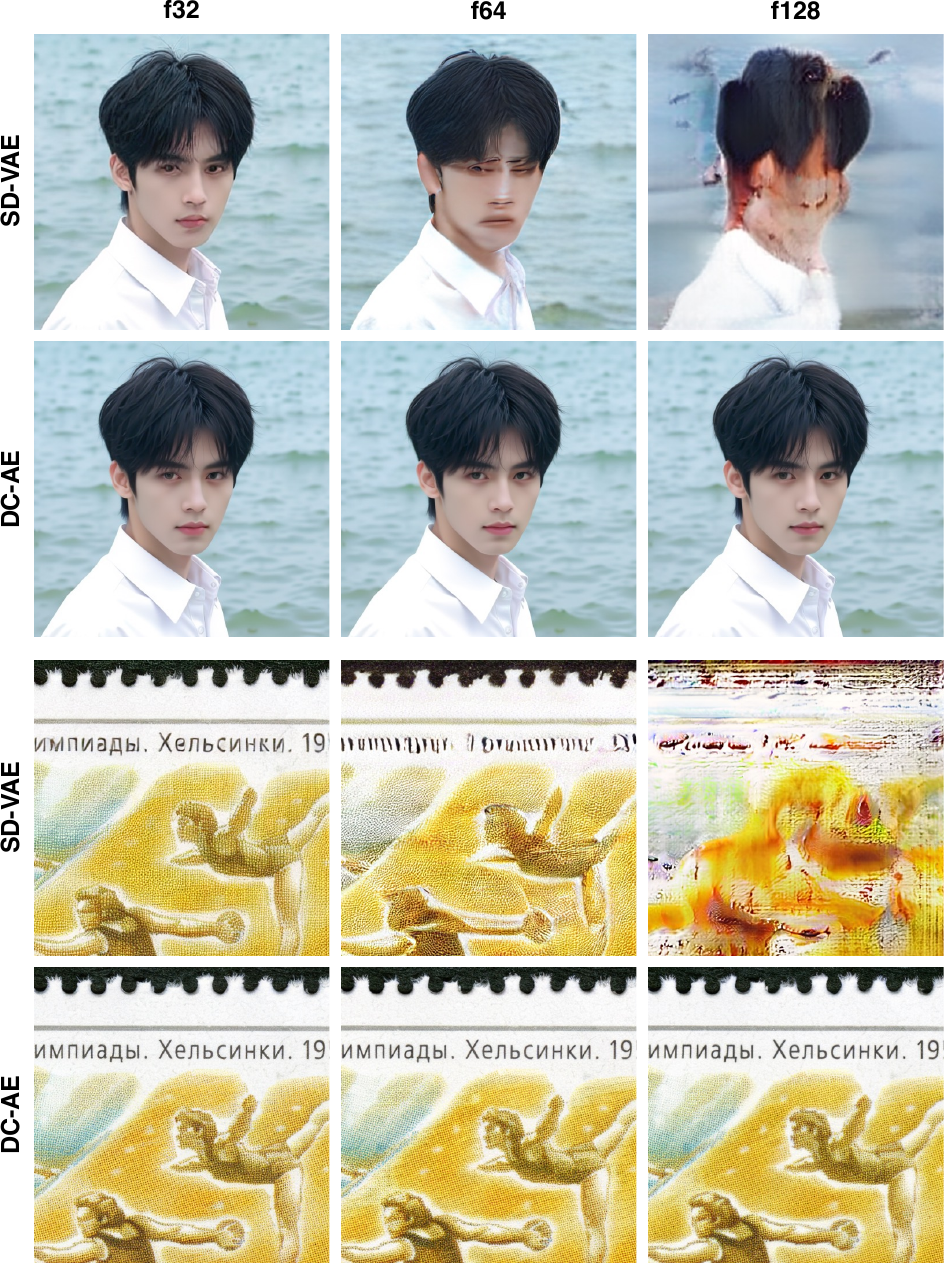}
    \caption{\textbf{Additional Autoencoder Image Reconstruction Samples.}}
    \label{fig:ae_visualization_1}
\end{figure}

\begin{figure}[ht]
    \centering
    \includegraphics[width=1\linewidth]{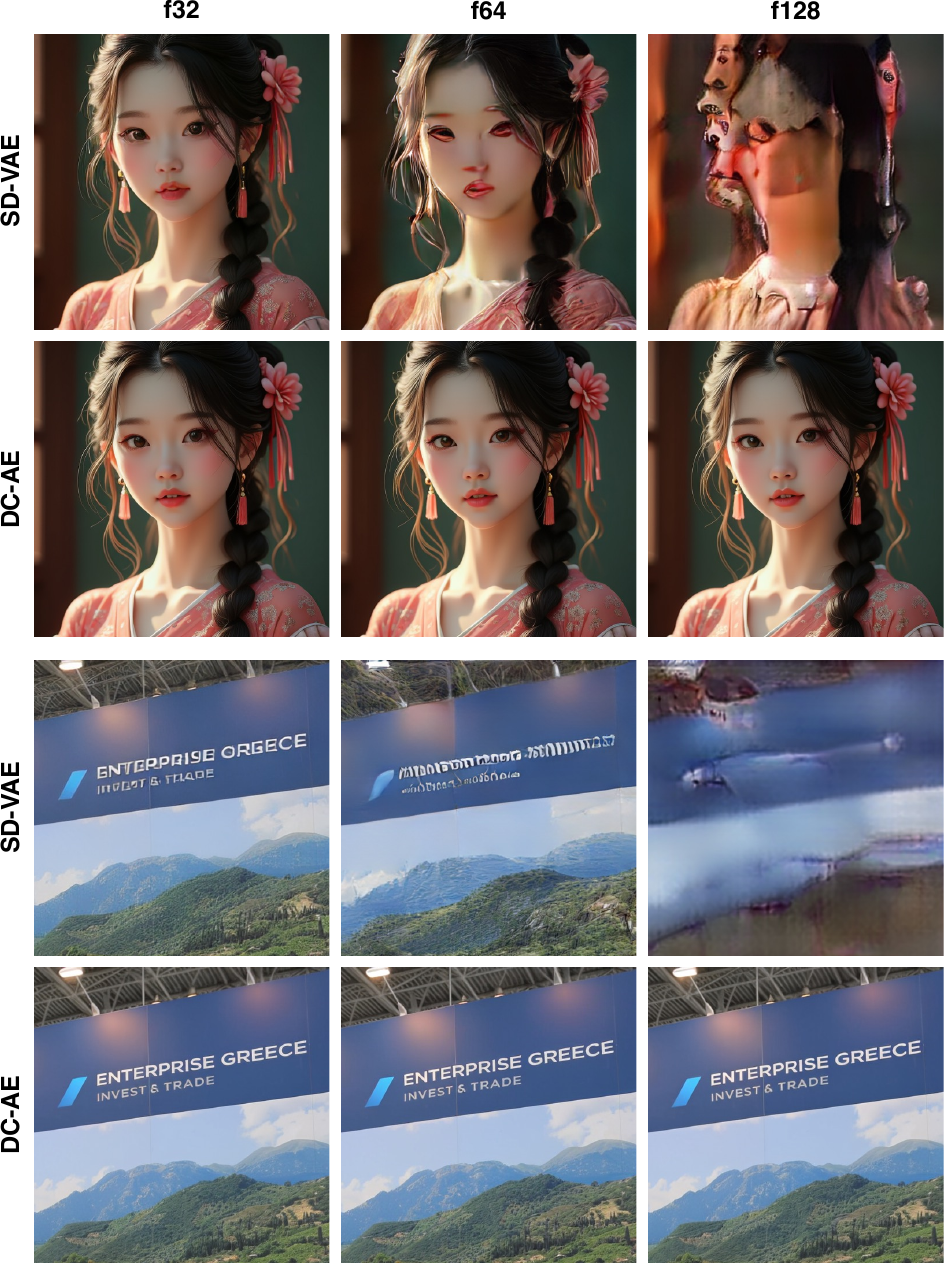}
    \caption{\textbf{Additional Autoencoder Image Reconstruction Samples.}}
    \label{fig:ae_visualization_2}
\end{figure}

\begin{figure}[t]
    \centering
    \includegraphics[width=1\linewidth]{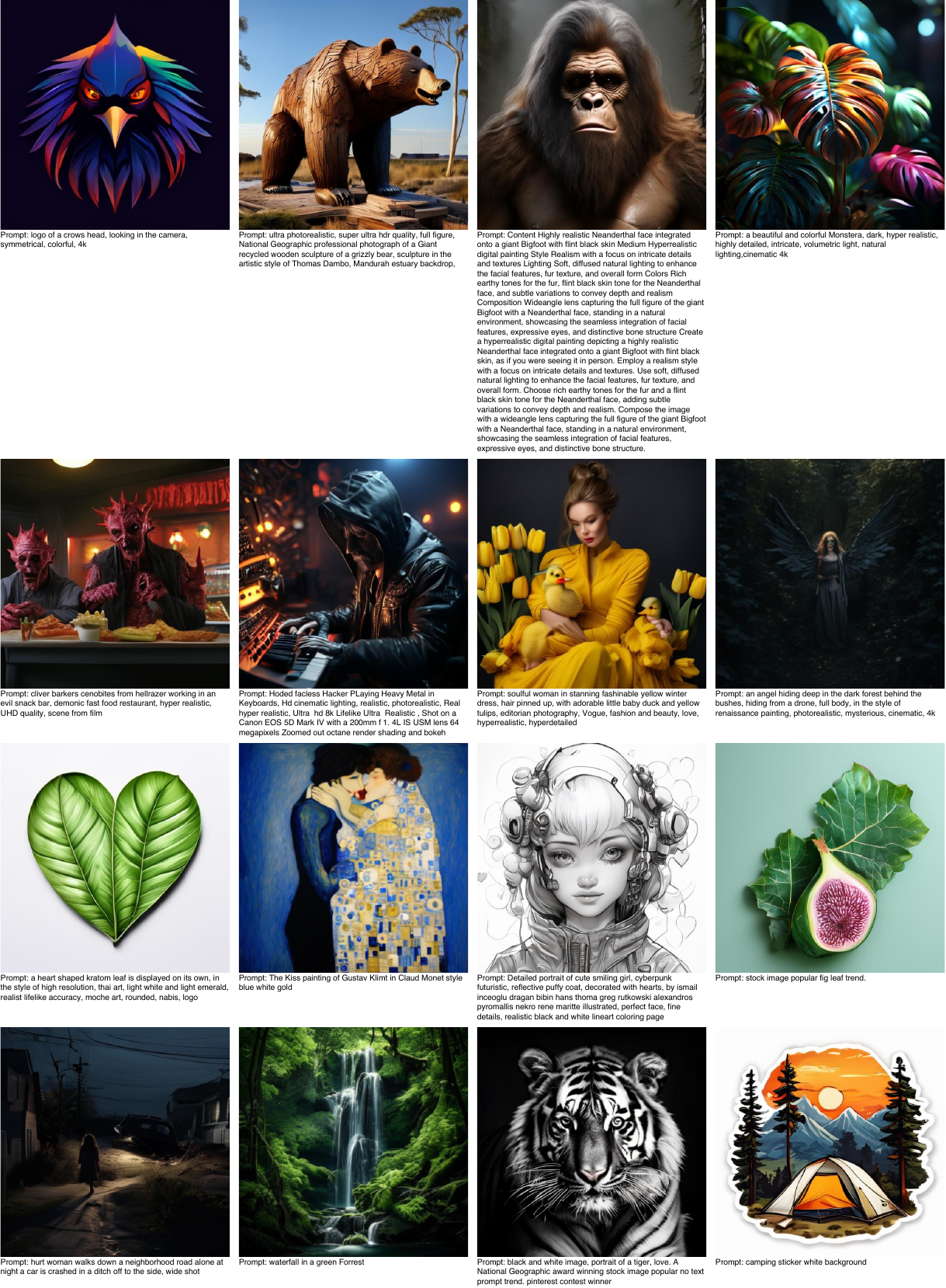}
    \caption{\textbf{Random 512$\times$512 Text-to-Image Samples.} Prompts are randomly drawn from MJHQ-30K \citep{li2024playground}.}
    \label{fig:diffusion_text_to_image_random}
\end{figure}

\begin{figure}[t]
    \centering
    \includegraphics[width=1\linewidth]{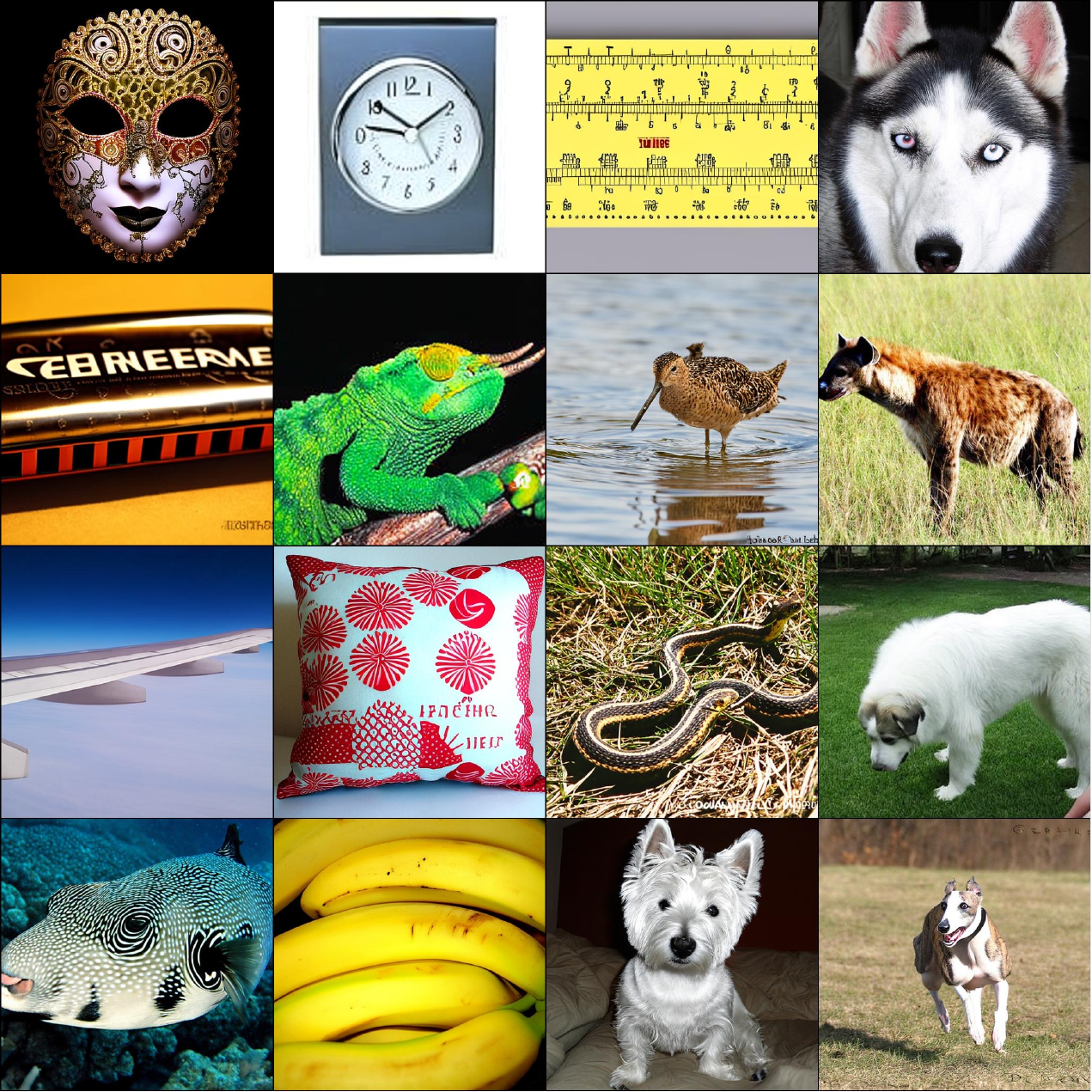}
    \caption{\textbf{Random Generated Samples on ImageNet 512$\times$512.}}
    \label{fig:diffusion_imagenet_random}
\end{figure}

In Figure \ref{fig:ae_visualization_1} and \ref{fig:ae_visualization_2}, we provide additional image reconstruction samples produced by SD-VAE and \modelshort. Reconstructed images by \modelshort demonstrate better visual qualities than SD-VAE’s reconstructed images, especially for the f64 and f128 autoencoders. Some samples are cropped for better visualization of details like human faces and small texts.

In Figure \ref{fig:diffusion_text_to_image_random} and Figure \ref{fig:diffusion_imagenet_random}, we show randomly generated samples on ImageNet 512$\times$512 and MJHQ-30K 512$\times$512 by the diffusion models using our \modelshort.

\end{document}